\documentclass{article}

\usepackage[preprint,nonatbib]{neurips_2025}
\usepackage[numbers]{natbib} 

\usepackage[utf8]{inputenc} 
\usepackage[T1]{fontenc}    
\usepackage{hyperref}       
\usepackage{url}            
\usepackage{booktabs}       
\usepackage{amsfonts}       
\usepackage{nicefrac}       
\usepackage{microtype}      

\usepackage[table]{xcolor}

\usepackage[utf8]{inputenc} 
\usepackage[T1]{fontenc}    
\usepackage{hyperref}       
\usepackage{url}            
\usepackage{booktabs}       
\usepackage{amsfonts}       
\usepackage{nicefrac}       
\usepackage{microtype}      

\usepackage{amsmath} 

\usepackage{multirow}

\usepackage{subcaption}

\usepackage{algpseudocode}
\usepackage{algorithm}

\usepackage{array} 
\usepackage{graphicx,multirow}

\usepackage{pifont}
\newcommand{\tick}{\ding{51}}
\newcommand{\cross}{\ding{55}}

\title{
Do Real-World Datasets Contain Natural Experiments? An Empirical Study Using \\ Causal Feature Selection 
}

\author{%
  Gautam Gare \\
  Carnegie Mellon University\\
  \And
  John Galeotti \\
  Carnegie Mellon University\\
  \And
  Michael Mozer \\
  Google DeepMind \\
  \AND
  Deva Ramanan \\
  Carnegie Mellon University\\
  \And
  Nan Rosemary Ke \\
  Google DeepMind \\
}

\begin{document}

\maketitle

\begin{abstract}

In nature, events that affect some individuals or groups but not others constitute an implicit intervention and are known as natural experiments. For example, the COVID-19 pandemic was an intervention by the coronavirus on the sub-population infected with COVID.  
We ask, \emph{do natural experiments occur in existing real-world datasets? If yes, how should we treat them?} To detect natural experiments in data, we use causal discovery to recover the underlying causal graph and perform feature selection based on causal links. If downstream performance improves by treating the data as interventional rather than observational, we argue that this suggests the dataset contains natural experiments. We first validate this hypothesis by simulating datasets with and without natural experiments using synthetic graphs. We then perform a systematic empirical evaluation on a large suite of real-world datasets.
Our results indicate that \emph{real-world datasets do contain natural experiments and we can take advantage of those natural experiments to improve model performance using causal inference}.
Our work represents the initial foray into this area, offering a preliminary exploration within a limited scope.
\end{abstract}

\section{Introduction}

Natural experiments are a type of research design used in social science, economics, and other fields where it is not feasible, ethical, or desirable to conduct a randomized controlled trial \citep{Molenberg2022NaturalChallenges}. In a natural experiment, researchers observe the effects of a naturally occurring event or intervention that affects some individuals or groups but not others. This provides an opportunity to make causal inferences about the effects of the event or intervention, assuming that other factors are held constant \citep{pearl2009causal,Gianicolo2020MethodsPublicationsFix}.
For example, a researcher might study the effects of a minimum wage increase by comparing employment and wage trends in states that raised the minimum wage with those that did not. Another example might be the study of the impact of a natural disaster, such as a hurricane, on economic or social outcomes.

One of the strengths of natural experiments is that they can provide a real-world context for studying the effects of an intervention or event, and may be more generalizable than findings from a laboratory experiment. 
Thus, the use of natural experiments has been recognized as a valuable tool in many fields of research, including those honored by the Nobel Prize. This methodology was heavily elevated in 2021, when David Card, Joshua Angrist, and Guido Imbens received the Nobel Prize in Economics \cite{Ball2021Nobel-winningRobust} for demonstrating that real-world policy shifts and accidental events can serve as rigorous, quasi-experimental frameworks to answer critical social questions.

However, natural experiments also have distinct limitations, such as the lack of control over the intervention or event being studied and the potential for other factors to confound the results. To address these nuances, 2019 Nobel laureates Abhijit Banerjee, Esther Duflo, and Michael Kremer \citep{2019UnderstandingAlleviation} view natural experiments as a useful complementary tool when active randomization is impossible. Ultimately, both methodologies enrich empirical economics: while field-based Randomized Controlled Trials (RCTs) maximize internal validity and replicability, natural experiments provide an invaluable window into real-world dynamics, allowing researchers to study vital social phenomena that would otherwise remain out of reach.


Our central hypothesis is as follows: \emph{do natural experiments exist within real-world ML datasets, and if so can we take advantage of them?}
Our experiments suggest a positive answer to both questions. 
By identifying natural experiments that may exist in real-world data, we can exploit causal models that lead to more accurate predictions and better decision-making. Our work represents the initial foray into this area, offering a preliminary exploration within a limited scope.

\begin{figure}[t!]
\centering
\includegraphics[width=.55\columnwidth]{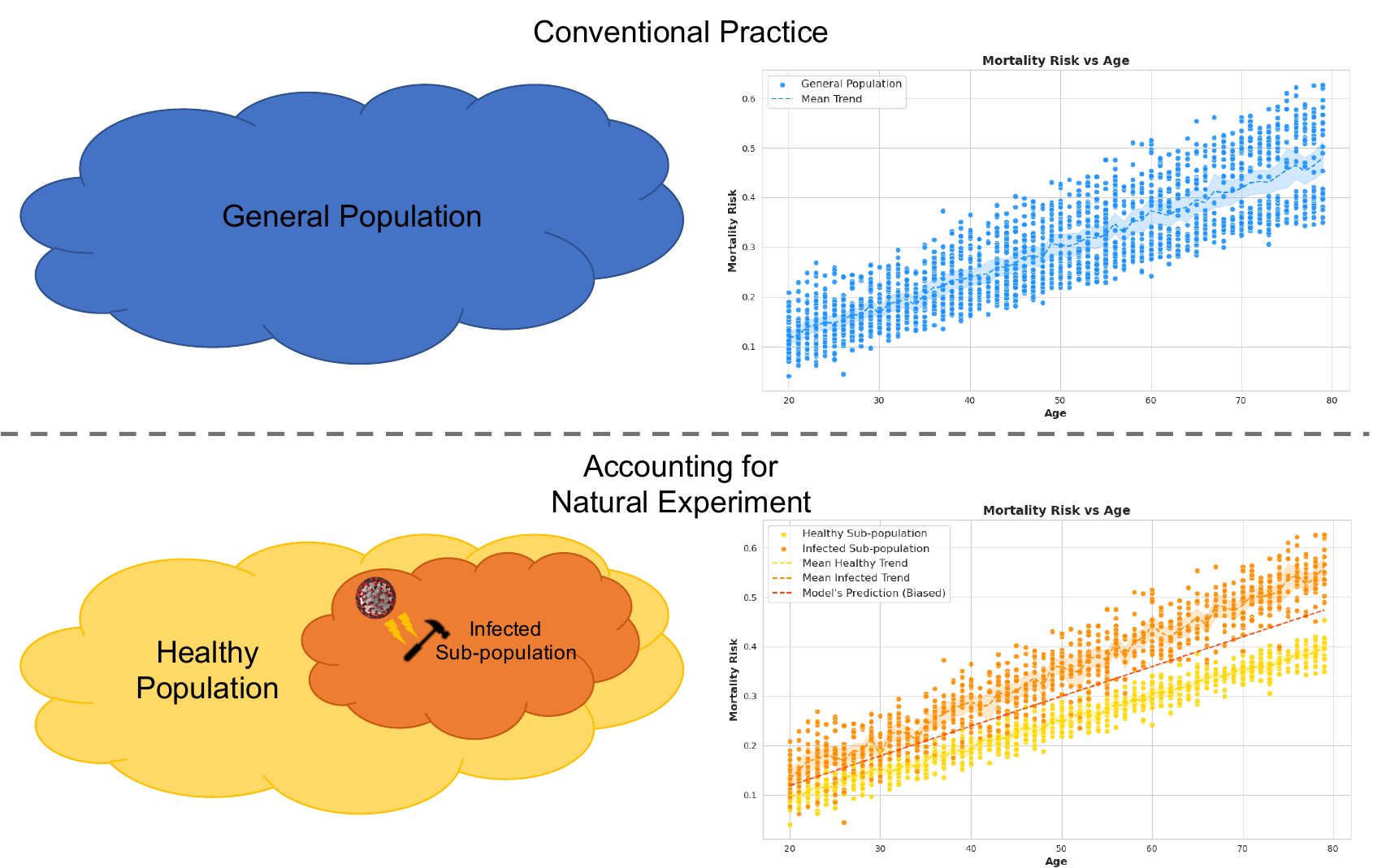}
\includegraphics[width=.4\columnwidth]{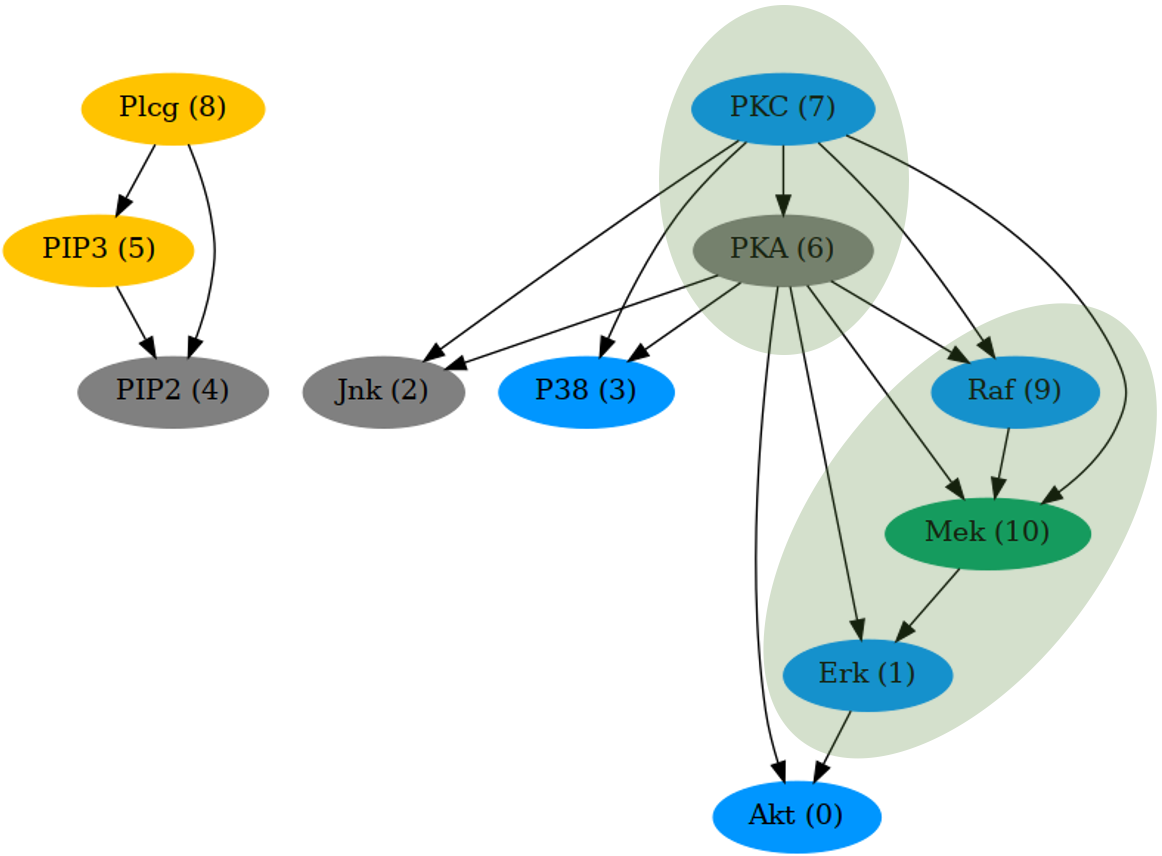}
\caption[(\textbf{Left}) Conventional vs Causal models. (\textbf{Right}) Sachs DAG graph.]{(\textbf{Left}) Conventional ML approaches do not account for naturally occurring experiments in data, which can result in biased predictions. Causal approaches which capture naturally-occuring interventions can lead to more accurate predictions; please see 
Appendix~\ref{app:simple_ex} 
for a simple illustrative demonstration. (\textbf{Right}) The causal graph used for synthetic Sachs \citep{sachs2005causal}  experiments, as in the BnLearn library \citep{Scutari2010LearningPackage}. In our experiments, we treat the Mek variable (the green node) as the target class to be predicted when measuring downstream classification accuracy. It's Markov-blanket (the blue region) includes PKC, PKA, Raf, and Erk. Hidden confounders (the grey nodes) are PKA, PIP2, and Jnk. 
The node numbers are shown in parentheses.
} 
\label{fig:teaser}
\end{figure}

{\bf Key idea.} We uncover natural experiments in real-world data by using causal discovery algorithms 
to infer the underlying causal graph, and use the graph to perform feature selection for downstream classification tasks. 
Specifically, we treat a sub-portion of the data (based on the classification label) to be interventional,
and then
we
run causal discovery algorithms  
to learn a model that generalizes to 
the entire dataset, including both the observational and interventional portions.
If downstream performance improves by treating the data as interventional rather than observational, we argue that this suggests the dataset contains natural experiments.
Importantly, we first validate this hypothesis in simulation using synthetic datasets with and without natural experiments. From this perspective, we show that synthetic datasets commonly used for causal analysis can {\em also} be used to emulate natural experiments by introducing interventional splits of training and test data, both with known and unknown interventional structure, and known and unknown hidden confounders. To our knowledge, such an analysis is novel. 
We use our synthetic experiments to validate our approach, and apply the most effective variant on a large suite of real-world datasets and find that a significant fraction (3 out of 11) may indeed be modeled as containing natural experiments.


\textbf{Contributions.} In this work, we propose a method to detect natural experiments in real-world datasets and demonstrate that we can train better models by taking natural experiments into account. We use the recent DCDI algorithm \citep{Brouillard2020DifferentiableData} to 
infer the underlying causal graph.
We use this graph for feature selection in a downstream classifier, by retaining only those features in the Markov blanket of the target variable to be classified. 
We validate our approach on eleven public real-world datasets, comparing it to popular feature selection baselines that are non-causal. We find that some \emph{real-world datasets do appear to contain natural experiments and we can take advantage of those natural experiments to train better downstream classifiers.} 

\section{Background}
Causal models are often described using Causal Bayesian Networks (CBNs). A CBN is a pair  $\mathcal{M}=\langle\mathcal{G}, p\rangle$, which defines a distribution $p$ over a set of random variables $\{X_1,\ldots, X_N\}$ and a \emph{directed acyclic graph} (DAG) $\mathcal{G}$, whose nodes represent the random variables and edges represent the causal dependencies among these variables. The joint distribution $p$  over all nodes is factorized into the product of the conditional probability distributions (CPD) of each node $X_n$ given its \emph{parents} $\text{pa}(X_n)$ (namely all nodes with an edge onto $X_n$), such that,
$p(X_1, \ldots, X_N) = \prod_{n=1}^N p(X_n \,|\, \text{pa}(X_n))$. Causal discovery techniques aim to uncover the underlying causal relations among the features that govern the data distribution. The goal is to discover the cause-effect relations rather than just statistical correlations between features \citep{pearl1995causal,peters2017elements}. The data can either come from an observational distribution or an interventional distribution. An \textbf{observational data} sample is obtained passively (as opposed to performing experiments) and is drawn from the joint distribution $p(X_1, \ldots, X_N)$. \textbf{Interventions} on a causal model are changes that force variables to take on a value (or a distribution) independent of other variables, thus the interventional samples are drawn from an altered joint distribution $p_I
(X_1, \ldots, X_N)$. There are several types of interventions: \textbf{Hard interventions}, the target node is set to a fixed value, severing its connections from its parents. \textbf{Soft interventions}, the conditional distribution is altered, retaining some influence from parents. \textbf{Known or Unknown} interventions, depending on whether the target nodes of the intervention are known or not \citep{Brouillard2020DifferentiableData}.

The \textbf{Markov blanket} (MB) of a node constitutes the minimal set of features that renders the node independent of the rest \citep{pearl2009probabilistic}. The MB includes parents, children, and spouses (parents of children) of the node under consideration, see Figure \ref{fig:teaser} (right) for an example. For classification, the MB of the class nodes has been theoretically proven to be the optimal feature subset for classification \citep{ks-tofs-96,Tsamardinos2003TowardsWrappers}.

\section{Related Works}
Our work relates to prior works on feature selection and causal discovery.

\textbf{Causal Discovery.} Most existing causal discovery \citep{Pearl2009CAUSALITY2009} methods fall under the categories of score-based \citep{Heckerman1995LearningData} and constraint-based techniques \citep{Spirtes2000CausationSearch.}. Constraint-based methods infer the underlying graph by analyzing the conditional independence in the data \citep{monti2019causal,spirtes2000causation,sun2007kernel,zhang2012kernel,Zhu2019CausalLearning}. Score-based techniques search through possible graphs (or DAGs) by assigning scores for graphs and ranking them \citep{chickering2002optimal,cooper1999causal,goudet2017causal,hauser2012characterization,heckerman1995learning,Tsamardinos2006TheAlgorithm,Huang2018GeneralizedDiscovery.,Zhu2019CausalLearning}. 
A recent surge of works that uses neural-networks for causal discovery have enabled scaling up to much larger graphs, while achieving state-of-art results on many datasets. These methods are score-based methods and they either frame causal discovery as a continuous optimization problem \citep{zheng2018dags,lee2019scaling,Yu2019DAG-GNN:Networks,lachapelle2019gradient,zheng2020learning,brouillard2020differentiable} or a mix of continuous and discrete optimization  \citep{bengio2019meta,KeLearningInterventions,Lippe2021EfficientConstraints}. In our work, we use DCDI \citep{Brouillard2020DifferentiableData} 
not only because it is one of the best performing method, but also for its compatibility with both interventional and observational data.

\textbf{Feature selection} methods can broadly be classified into filter, embedded, and wrapper methods. Filter methods are preferable as they are model-agnostic, unlike the other two \citep{Li2016FeaturePerspective,Yu2018ASelection}. The filter selection methods rank features based on relevance \citep{Yu2018ASelection}. However, unlike causal feature selection methods, the exact number of relevant features in filter methods is often indeterminate and must be prespecified by the user or determined iteratively. This is a limitation compared to causal feature selection, which identifies a fixed set of relevant features specified by the Markov blanket.

\textbf{Causal (non-deep) feature selection} algorithms \citep{Yu2019Causality-basedEvaluations,Yu2018ASelection} focus on recovering the Markov blanket rather than the top-k relevant features. \cite{pearl2009probabilistic} shows the Markov-blanket defines a sufficient set of relevant features, where the Markov-blanket is defined on the data governing causal graph. \cite{Yu2018ASelection} established the relations between classical (non-causal) and causal feature selection method objectives.
Most causal feature selection methods use conditional-independence tests to recover the Markov blanket (MB) or the partial Markov blanket (parents and children, PC) without inferring the entire causal graph, due to scalability issues. In contrast, we employ neural causal discovery methods that scale well with the number of features and data samples \citep{Brouillard2020DifferentiableData, Lippe2021EfficientConstraints}, allowing us to discover the entire causal graph on datasets with up to 50 features and 284,000 samples. The complete graph can provide insights into features and support applications like synthetic data generation.

\section{Methodology}
Our algorithm involves three steps, 1) discover the causal graph 2) determine the Markov-blanket (MB) for the target variable of interest 3) train the classifier using only the features in Markov Blanket.

\subsection{Differentiable Causal Discovery}
Recent neural causal discovery methods make use of differentiable continuous optimization to infer causal structures~\citep{zheng2018dags,ke2019learning,lee2019scaling,brouillard2020differentiable,lippe2021efficient,Ke2022LearningStructure}. Many of these methods can handle high-dimensional non-linear data and scale well with the number of nodes and data samples, making them suitable for real-world settings. 

We use DCDI \citep{Brouillard2020DifferentiableData} to learn the causal graph structure that models the data distribution. DCDI 
supports both observational and interventional data, including soft and hard interventions, as well as known and unknown interventions. For synthetic data experiments, we run DCDI under observational $O$ and interventional settings: known hard intervention $I_{HK}$, known soft intervention $I_{SK}$, and unknown hard intervention $I_{HU}$.\footnote{DCDI does not support unknown soft intervention setting.}

DCDI formulates the discrete search problem of discovering the causal graph as a continuous optimization. For a dataset with $N$ variables, the model consists of $N$ MLPs, each modeling the conditionals of one variable given a subset of the other $N-1$ variables. It learns a DAG-structured $N\times N$ adjacency matrix that represents the structure of the underlying causal graph. 
To learn the causal relationship between features $X$ and classification targets $Y$, we apply DCDI to their union $V = X \cup Y$, and treat the inferred DAG $G(V,E)$ with edge structure $E$ as the inferred causal model.

In the intervention setting, DCDI uses the same MLPs to model the intervention data as those used for the observational data, except for the target node of the intervention, which is modeled with a different MLP. This ensures that the DAG structure remains the same for both observational and interventional data, differing only at the target node of the intervention.

\subsection{Detecting natural experiments}
We argue that many real-world datasets can be taught of as having an interventional structure (natural experiments).  
For instance, COVID-19 classification data has an observational plus interventional data structure, where the healthy patient subset constitutes the observed (control) population and the sick patient subset constitutes the intervened (treatment) population, wherein the coronavirus is the intervening agent. This follows from the reasoning that a healthy person's lung has 
a natural functioning mechanism (which can be represented by a causal DAG), which when disturbed (intervened) by an external agent (virus) gives rise to a disease. Similarly, credit card fraud data may be viewed as having an interventional structure, where the fraudster alters the normal spending habits of a user. Many medical datasets and some non-medical real-world datasets can be interpreted in this way. 

Because real-world data tends to be noisy, we posit that most interventions should be treated as soft. 
We make the strong assumption that classification targets represent interventions, since they capture variables of interest (as motivated by the COVID-19 and credit fraud examples above). This allows us to treat our real-world datasets as containing soft and known interventions. Though features can also be intervened, this is beyond the scope of this paper. 

To evaluate our hypothesis, we run the causal discovery algorithm using the soft-known configuration, partitioning the data into observational and interventional based on class labels. In multi-class classification, we treat class-0 as observational and every other class as different interventional data. If downstream performance improves by treating the data as interventional rather than observational, we argue that this suggests the dataset contains natural experiments. 
We begin by assessing this mechanism on synthetic data generated from a known graph, facilitating precise measurement of the discovered graph's accuracy. We then proceed to testing on real-world datasets.
%

\textbf{Feature selection.}
Markov-Blanket (MB) constitutes the smallest subset of features sufficient for classification \citep{pearl2009probabilistic,Yu2019Causality-basedEvaluations}. The Markov-Blanket includes parents, children, and spouses (parents of children) of the node under consideration. We take the union of the MB of all the class nodes in the discovered causal graph, this now constitutes the causal features that will be used for classification. 

\textbf{Classification.}
To classify $x \in R^n$ into label $y$ we train a network that models the conditional probability $p(y|x) = p(y|F(x))$, where $F(x) = \{x_i: x_i \in MB(y)\}$.

\section{Experiments}
Our experiments first assess DCDI's ability to accurately recover the underlying causal graph from synthetic data generated from known causal graphs under various conditions (observational and/or interventional) akin to real-world scenarios and compare it against several baseline approaches. We then apply DCDI to real-world data, specifically evaluating its effectiveness in feature selection on popular benchmarking tabular datasets in the literature.

\paragraph{Hyperparameters.}
For all experiments, we search for the best MLP hyperparameters per dataset using Bayesian hyperparameter optimization \citep{Shahriari2016TakingOptimization} 
(refer Appendix Table~\ref{tab:hyperparams} for search space). 
The objective is to optimize the F1-score on the validation dataset. We use the same MLP hyperparameters with a fixed seed for all classifiers to ensure a fair comparison between methods. 
The DCDI  \href{https://github.com/slachapelle/dcdi}{codebase} provided by the authors is used for experiments, with default hyperparameters except for the learning rate and the weight of the coefficient $\lambda$.
See 
Appendix Table~\ref{tab:hyperparams} 
for more details. Our models are trained on Nvidia RTX A6000 GPU using AdamW \citep{Loshchilov2017DecoupledRegularization} optimizer. Runs that discover cyclic graphs are ignored, and we report classification performance using multi-layer perceptron (MLP) classifiers.
%

\paragraph{Feature Selection Baselines.}
We compare to the following classic feature selection baselines: 
a) Pearson's r co-efficient \citep{Bishara2012TestingApproaches} b) Spearman correlation \citep{Bishara2012TestingApproaches}  c) MIM (mutual information maximization) \citep{Lewis1992FeatureCategorization} d) Anova (ANalysis Of VAriance) F1-test \citep{Johnson2002PatternAnalysis} e) Chi-square \citep{HuanLiuChi2:Attributes}. These measures rank the features from which top-ranking features are used for classification. We perform a grid search on k in the range $[0.1, 0.9]$, where k is the ratio of selected features to all features. We also compare against Boruta \citep{Kursa2010BorutaSelection} which automatically determines the optimal number of features using a classifier (RandomForest with 100 trees of max depth 50). 

\paragraph{(Non-neural) Causal Feature Selection Baselines.} We also compare to (non-neural) causal feature-selection baselines: a) Lessen swamping, resist masking and highlight the true positives (LRH) \citep{Liu2016SwampingDiscovery} b) Simultaneous MB discovery (STMB) \citep{Gao2017EfficientApplication.} c) HITON-MB \citep{Aliferis2003HITON:Selection.} d) Min-Max Markov Blanket (MMMB) \citep{Tsamardinos2006TheAlgorithm} e) Incremental Association-Based Markov Blanket (IAMB) \citep{Tsamardinos2003AlgorithmsDiscovery}. 
These methods locally discover the Markov-blanket of the target node by efficiently searching through the nodes using conditional independence test. 
The tests are conducted using Fisher Z-test or $\chi^2$ test thresholded at a $\alpha$ significance level. We perform a grid search on $\alpha$ in the range $[0.01, 2]$.
%

\paragraph{Metrics.}
 When the ground-truth graph is available, we report the Structural Hamming Distance ($SHD$) between the discovered and ground-truth graphs. The discovered Markov blanket (MB) is also reported using the model with the highest F1-score on the test set, instead of the model with the lowest $SHD$. This approach mirrors real-world scenarios where the ground-truth causal graph is unknown. Additionally, we measure the edit distance ($ED_{MB}$) compared to the ground truth. Finally, we report the F1-score for the classifiers.

\subsection{Synthetic Data Experiments}
\label{sec:synthetic}
We evaluate causal discovery performance of DCDI on the widely-used synthetic Sachs \citep{sachs2005causal} graph from the Bayesian Network Repository (BnLearn) \citep{Scutari2010LearningPackage}.
Sachs has 11 nodes and 17 connected edges (see Figure~\ref{fig:teaser}). To transform this graph into a classification problem, we select the `Mek' node as the target node to be classified. This node is chosen because it is not a source or a leaf node, and it has both ancestors and descendants, which is typical of nodes in real-world datasets.
The `Mek' node is a ternary variable, making it a 3-way classification problem.
To generate data, we follow established strategies from previous works \citep{ke2019learning,lippe2021efficient,Ke2022LearningStructure}. We generate $10$k observational data points, which are then split into train, val, and test sets (70:10:20 split). For intervention data, we generate $1$k intervention samples per node, which are used along with the train set for the intervention experiments. 

Real-world datasets can be either solely observational or observational plus interventional, where algorithms may either be aware or unaware of the intervened samples. We emulate such settings with our synthetic data: 1) Interventions on all variables 2) Interventions on few variables 3) Observational data setting 4) Observational data with hidden confounders. 

\begin{table}[h]
 \caption[Causal feature selection results on the synthetic Sachs data]{
      Causal feature selection on the synthetic Sachs data~\citep{Scutari2010LearningPackage}, with different types of data interventions and varying influence of hidden confounders.  We analyze causal algorithms with various knowledge of these interventions ($O$, $I_{SK}$, $I_{HU}$, $I_{HK}$) and feature selection baselines (chi2,...boruta), comparing their ability to recover the true Markov-blanket, measured in terms of edit distance ($ED_{MB}$). Across all settings, \emph{causal algorithms outperform feature selection baselines, both in terms of recovering truly relevant features in the Markov-blanket and in classification accuracy}. 
       {\bf To ensure a fair comparison, we fix random seeds when training classifiers, ensuring methods that discover the same Markov-blanket will have identical accuracy.}
       Causal $O$ and $I_{SK}$ perform consistently well, even for purely observational data and observational data with hidden confounders. More details are in Section~\ref{sec:synthetic}. Though most methods have comparable F1-scores, causal methods most accurately discover and use causal (Markov blanket) features, making them robust to shifts.
      }  
  \label{tab:synthetic_cls_exp}
 \begin{minipage}{.5\linewidth}
\resizebox{\columnwidth}{!}{
  \begin{tabular}{l|cc|c|cc}
    \toprule
    
    Method & \multicolumn{2}{c}{F1-score $\uparrow$} & \textbf{Markov-blanket} & $SHD_G \downarrow$ & $\bf{ED_{MB} \downarrow}$ \\ 
     & obs+interv & obs & node-number & & \\ 

    \midrule
    \midrule
    \multicolumn{6}{c}{Observation + Intervention on all variables} \\  
    \midrule
    
    All-features & - & 0.79 & all $[0, 9]$ & - & -  \\
    Ground truth & - & 0.79 & 1, 6, 7, 9 & 0 & 0  \\
    \midrule
    
    \textbf{Causal $O$} & 0.79 & 0.79 & 1, 5, 6, 7, 9 & 17  & 1 \\
    \textbf{Causal $I_{SK}$} & - & 0.79 & 1, 6, 7, 9 & 5  & \textcolor{blue}{\textbf{0}} \\
    Causal $I_{HK}$ & - & 0.79 & 1, 6, 7, 9 & 7  & \textcolor{blue}{\textbf{0}} \\
    Causal $I_{HU}$ & - & 0.79 & 1, 5, 6, 7, 9 & 14 & 1 \\
    
    \midrule

    chi2 & 0.75 & 0.74 & 0, 3, 7, 9 & - & 4  \\
    spearman & 0.74 & 0.75 & 0, 1, 3, 9 & - & 4 \\
    mi & 0.77 & 0.76 & 0, 6, 7, 9 & - & 2 \\
    anova & 0.77 & 0.77 & 0, 1, 6, 9 & - & 2 \\
    pearson & 0.74 & 0.75 & 0, 1, 3, 9 & - & 4 \\
    boruta & 0.79  & 0.79 & 0, 1, 6, 7, 9 & - & 1 \\
    
    \midrule
    \midrule
    \multicolumn{6}{c}{Observation + Intervention on few variables [Mek, PKA, PIP3] (10, 6, 5)} \\  
    \midrule
    
    All-features & - & 0.79 & all $[0, 9]$ & - & - \\
    Ground truth & - & 0.79 & 1, 6, 7, 9 & 0 & 0 \\
    
    \midrule
    
    \textbf{Causal $O$} & 0.79 & 0.79 & 1, 5, 6, 7, 9 & 17 & 1 \\
    \textbf{Causal $I_{SK}$} & - & 0.79 & 1, 6, 7, 9 & 8 & \textcolor{blue}{\textbf{0}} \\
    Causal $I_{HK}$ & - & 0.79 & 1, 6, 7, 9 & 7 & \textcolor{blue}{\textbf{0}} \\
    Causal $I_{HU}$ & - & 0.79 & 1, 6, 7, 9 & 10 & \textcolor{blue}{\textbf{0}} \\
    
    \midrule

    chi2 & 0.75 & 0.75 & 0, 3, 6, 9 & - & 4 \\
    spearman & 0.75 & 0.75 & 0, 1, 3, 9 & - & 4 \\
    mi & 0.76 & 0.76 & 0, 6, 7, 9 & - & 2 \\
    anova & 0.75 & 0.75 & 0, 3, 6, 9 & - & 4 \\
    pearson & 0.75 & 0.75 & 0, 1, 3, 9 & - & 4 \\
    boruta & 0.79 & 0.79 & 0, 1, 6, 7, 9 & - & 1 \\

    \bottomrule
  \end{tabular}
    }
\end{minipage}
    \begin{minipage}{.5\linewidth}
  \resizebox{\columnwidth}{!}{
  \begin{tabular}{l|cc|c|cc}
    \toprule
    
    Method & \multicolumn{2}{c}{F1-score $\uparrow$} & \textbf{Markov-blanket} & $SHD_G \downarrow$ & $\bf{ED_{MB} \downarrow}$ \\ 
     & obs+interv & obs & node-number & & \\ 

    \midrule
    \midrule
    \multicolumn{6}{c}{Observational data} \\  
    \midrule

    All-features & - & 0.79 & all $[0, 9]$ & - & - \\
    Ground truth & - & 0.79 & 1, 6, 7, 9 & 0 & 0 \\
    
    \midrule
    \textbf{Causal $O$} & - & 0.79 & 1, 6, 7, 9 & 8 & \textcolor{blue}{\textbf{0}} \\
    \textbf{Causal $I_{SK}$} & - & 0.79 & 1, 5, 6, 7, 9 & 18 & 1 \\
    Causal $I_{HK}$ & - & 0.79 & 0, 1, 3, 5, 6, 7, 9 & 21 & 3 \\
    Causal $I_{HU}$ & - & 0.79 & 0, 1, 5, 6, 7, 9 & 8 & 2 \\

   \midrule

    chi2 & - & 0.75 & 0, 3, 6, 9 & - & 4 \\
    spearman & - & 0.75 & 0, 1, 3, 9 & - & 4 \\
    mi & - & 0.76 & 0, 6, 7, 9 & - & 2 \\
    anova & - & 0.75 & 0, 3, 6, 9 & - & 4 \\
    pearson & - & 0.75 & 0, 1, 3, 9 & - & 4 \\
    boruta & - & 0.79 & 0, 1, 3, 6, 7, 9 & - & 2 \\

    \midrule
    \midrule
    \multicolumn{6}{c}{Observation with hidden confounders [PKA, PIP2, Jnk] (6, 4, 2) } \\  
    \midrule
    
    All-features & - &  0.77 & all $[0, 9] \notin {2,4,6}$ & - & - \\
    Ground truth & - & 0.77 & 0, 1, 3, 7, 9 & 0 & 0 \\

    \midrule
    
    \textbf{Causal $O$} & - & 0.77 & 0, 1, 3, 7, 9 & 17 & \textcolor{blue}{\textbf{0}} \\
    \textbf{Causal $I_{SK}$} & - & 0.77 & 0, 1, 3, 7, 9 & 20 & \textcolor{blue}{\textbf{0}} \\
    Causal $I_{HK}$ & - & 0.77 & 0, 1, 3, 5, 7, 9 & 18 & 1 \\
    Causal $I_{HU}$ & - & 0.77 & 0, 1, 3, 5, 7, 9 & 19 & 1 \\

   \midrule

    chi2 & - & 0.77 & 0, 1, 3, 7, 9 & - & \textcolor{blue}{\textbf{0}} \\
    spearman & - & 0.75 & 0, 1, 3, 5, 9 & - & 2 \\
    mi & - & 0.77 & 0, 1, 3, 7, 9 & - & \textcolor{blue}{\textbf{0}} \\
    anova & - & 0.77 & 0, 1, 3, 7, 9 & - & \textcolor{blue}{\textbf{0}} \\
    pearson & - & 0.75 & 0, 1, 3, 5, 9 & - & 2 \\
    boruta & - & 0.77 & 0, 1, 3, 7, 9 & - & \textcolor{blue}{\textbf{0}} \\

    \bottomrule
  \end{tabular}
  } 
    \end{minipage}%
\end{table}

\subsubsection{Intervention on all variables}
\label{sec:interv_on_all}
In this setup, we have both observational and interventional data on all variables, which is ideal for causal discovery. Neural causal discovery techniques are theoretically guaranteed to recover the complete causal graph under this regime. 
We run DCDI with three intervention configurations: 1) hard and known intervention ($I_{HK}$), 2) hard and unknown intervention ($I_{HU}$), and 3) soft and known intervention ($I_{SU}$). The hard and known intervention should ideally give the best results, as it matches the test setup. As expected, DCDI successfully recovers the correct Markov blanket (MB) in both $I_{HK}$ and $I_{SK}$ setting (refer to Table~\ref{tab:synthetic_cls_exp}). For brevity, MB is stated in node numbers (cf. Fig~\ref{fig:teaser}).

\textbf{Treating interventions as observations} to simulate the scenario where the algorithm is unaware of the intervened data subset. We run DCDI using the observational ($O$) configuration on the combined observation and intervention data (obs+interv), and we find that it recovers the MB with one false positive. We report the F1-score of the classifier trained on obs+interv as well as solely on the observation data (obs) for comparison with DCDI intervention runs. In both cases, the F1-score is reported on the same test set of the observation data.  We also test the classical causal feature-selection methods by selecting the top-4 ranked features (as the ground-truth MB size is 4). These methods are unable to recover the exact MB, with Boruta performing the best with one false positive. Interventions (perfect) cause sparsification of the graph which makes correlated variables appear uncorrelated on the intervened data. 
So non-causal feature selection may down-weight such features.

\subsubsection{Intervention on few variables}
We now conduct experiments with interventions on only three variables (Mek, PKA, PIP3) and find that DCDI successfully discovers the Markov blanket (MB) under all configurations.

\textbf{Treating interventions as observations.} 
DCDI with observational settings again outperforms non-causal baselines, reliably recovering the MB (with one false positive). Causal methods are more effective on intervention data, even when the intervened samples are unknown ($I_{SU}$).

\subsubsection{Observational data}
Under the observation setting, we don't have any interventional data, so DCID with observation ($O$) is expected to perform better. As expected it reliably recovers the MB. The non-causal baselines again are unable to recover the correct MB even on the observational data.

\textbf{Treating observations as interventions.} We partition the data according to the Mek class labels, treating class-0 as observational data and the remaining two classes as interventional data. We then run DCDI using its intervention-aware configuration. Although this setup leaves only the class-0 samples contributing to the observational likelihood objective for Mek node, which could potentially lead to degeneracy during optimization, we do not observe such issues in practice and obtain stable convergence. Table~\ref{tab:synthetic_cls_exp} shows that the soft-known intervention configuration, $I_{SK}$, performs competitively, yielding only a single false positive.

\subsubsection{Observational data with confounders}
Now, we consider the scenario when hidden confounders are present. We simulate this by marginalizing out nodes PKA, PIP2, and Jnk such that they are not observed in the data. In the resulting Marginalized Mixed Acyclic Graph (MAG) \citep{Liu2021LearningEdges} 
(refer to Appendix Fig~\ref{fig:sachs_skip_graph}),
the Markov blanket (MB) comprises Akt, Erk, P38, PKC, and Raf. The $SHD_G$ metric is reported on this marginalized MAG graph.

Even in this setup, both DCDI ($O$ and $I_{SK}$) recover the exact MB. The non-causal baselines too recover the exact MB (except spearman and person). However, it must be noted that this is an easier setup as the methods need to select 5 out of 7 features.

\begin{table}[h]
\caption[({\bf Left}) Real world dataset statistics. ({\bf Right}) Synthetic data test set results.]{
  ({\bf Left}) Real world dataset statistics. ({\bf Right}) Synthetic data experiment with test data as observational ($O$), interventional ($I$), or both ($O+I$). We find causal interventional method (Causal $I_{SK}$) performs best when natural experiments are present, motivating further exploration on real-world data. Interventional data includes 1k samples per variable for PKA and PIP3.
      }  
\resizebox{\columnwidth}{!}{
 \begin{tabular}{|l|cccc|}
    \toprule
    Dataset     & Feature type      & \# of samples  & \# of features & \# of classes \\
    
    \midrule
    
    Adult   &  mixed & 48842  &  14  & 2 \\
    Buddy   & mixed  & 18834  &  9  & 3 \\
    Cardio   &  mixed & 70000  &  11  & 2 \\
    Churn Modelling   & mixed  & 10000  & 11   & 2 \\
    Credit Card Fraud   & mixed  &  284807 &  30  & 2 \\
    Dermatology   & numeric  & 358  &  34  & 6 \\
    Diabetes   &  mixed &  768 & 8   & 2 \\
    Gesture Phase   &  numeric & 9873  & 32   & 5 \\
    Higgs Small   &  numeric & 98049  & 28   & 2 \\
    Miniboone   &  numeric & 130064  & 50   & 2 \\
    Wilt   &  numeric & 4839  & 5   & 2 \\
        
    \bottomrule
  \end{tabular}
  \begin{tabular}{|l|ccc|}
    \toprule

     \multirow{2}{*}{Method} &  \multicolumn{3}{c}{Test set F1-score}\\
      & $O$ & $I$ & $O+I$ \\

    \hline
    Causal $O$ & \textbf{0.795} & 0.844 & 0.820 \\
    Causal $I_{SK}$ & 0.793 & \textbf{0.848} & \textbf{0.822} \\
    \midrule
    Natural Exp  & \cross & \tick & \tick \\
    \bottomrule
    \end{tabular}
  } 
\label{tab:stats}
\end{table}


\textbf{Findings.}
We observe that soft-known intervention ($I_{SK}$) configuration performs well across all four data settings. 
Since the true nature of real-world data is unclear (either observational or partly interventional), we propose evaluating DCDI using both $I_{O}$ and $I_{SK}$ configurations on these datasets.

\subsubsection{Interventional and Observational test data}
As the nature of the test set is also unclear on real-world datasets, we simulated observational and interventional test set combinations on the synthetic Sachs dataset and evaluated models trained on the ``Intervention on few variables'' setting (which closely resembles real-world datasets). The different test dataset variants are: observational ($O$), interventional ($I$), or a combination of both ($O+I$) data types. We find (cf. Table~\ref{tab:stats} right) causal observational setting (Causal $O$) performs better on observational ($O$) test sets, while the causal interventional setting (Causal $I_{SK}$ ) excels when interventional data is present in the test set ($I$, $O+I$). This suggests that Causal $I_{SK}$ can be used as an indicator of natural experiments (interventional data) in the dataset whenever it outperforms other methods, thus validating our hypothesis. We now apply this hypothesis to real-world datasets.

\newlength{\width}
\setlength{\width}{0.95 in}
\newlength{\height}
\setlength{\height}{0.70 in}

\begin{figure*}[!ht]
\centering

\setlength{\tabcolsep}{0.01pt} 
\def\arraystretch{0.01} 

\newcolumntype{C}{>{\centering\arraybackslash}m{\width}<{}}
\newcolumntype{F}{>{\centering\arraybackslash}m{0.35\width}<{}}
\resizebox{\textwidth}{!}{
\begin{tabular}{CCC}

\subfloat{\includegraphics[height = 0.75\height, width = \width]{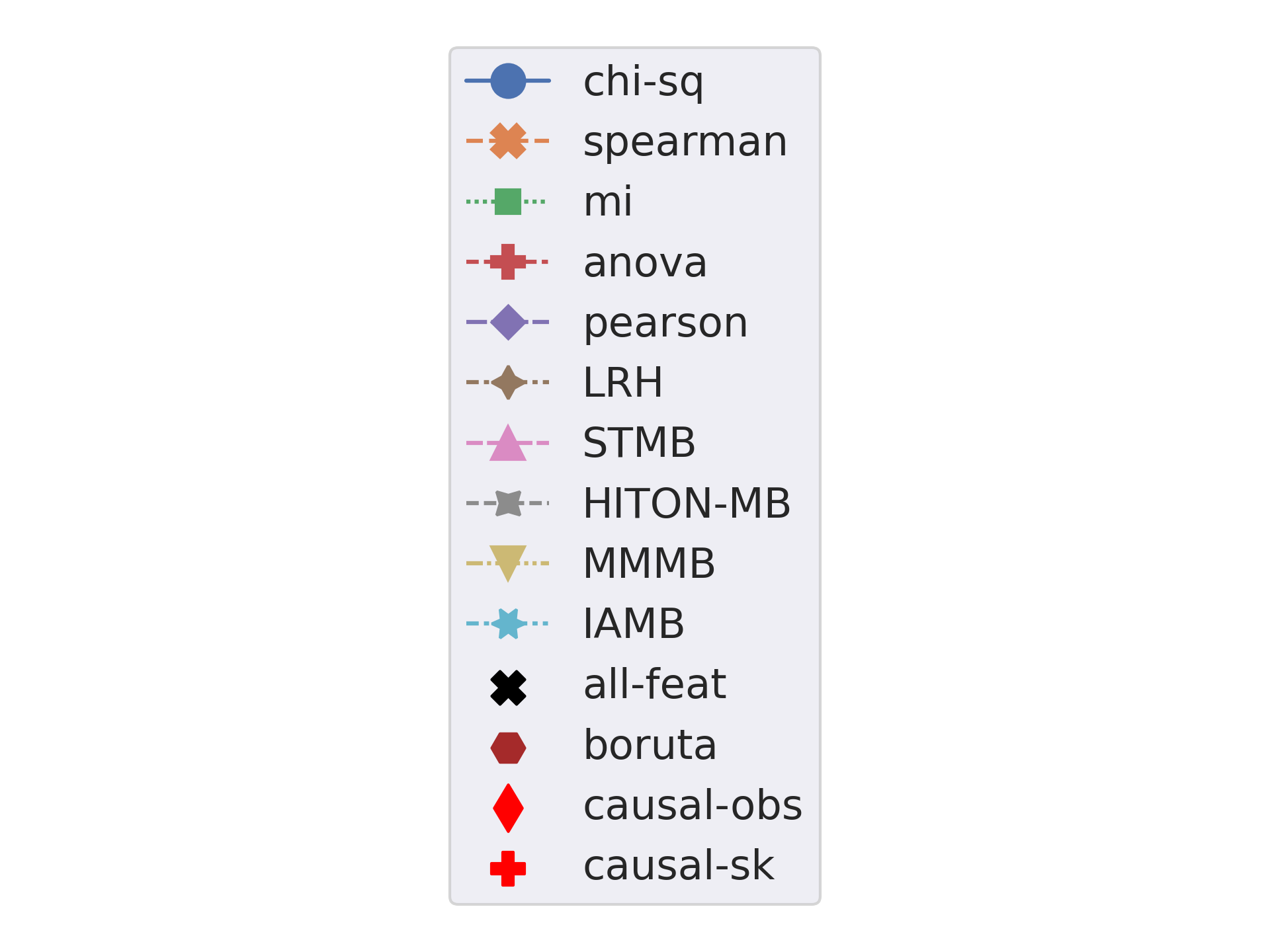}} &
\subfloat{\includegraphics[height = \height, width = \width]{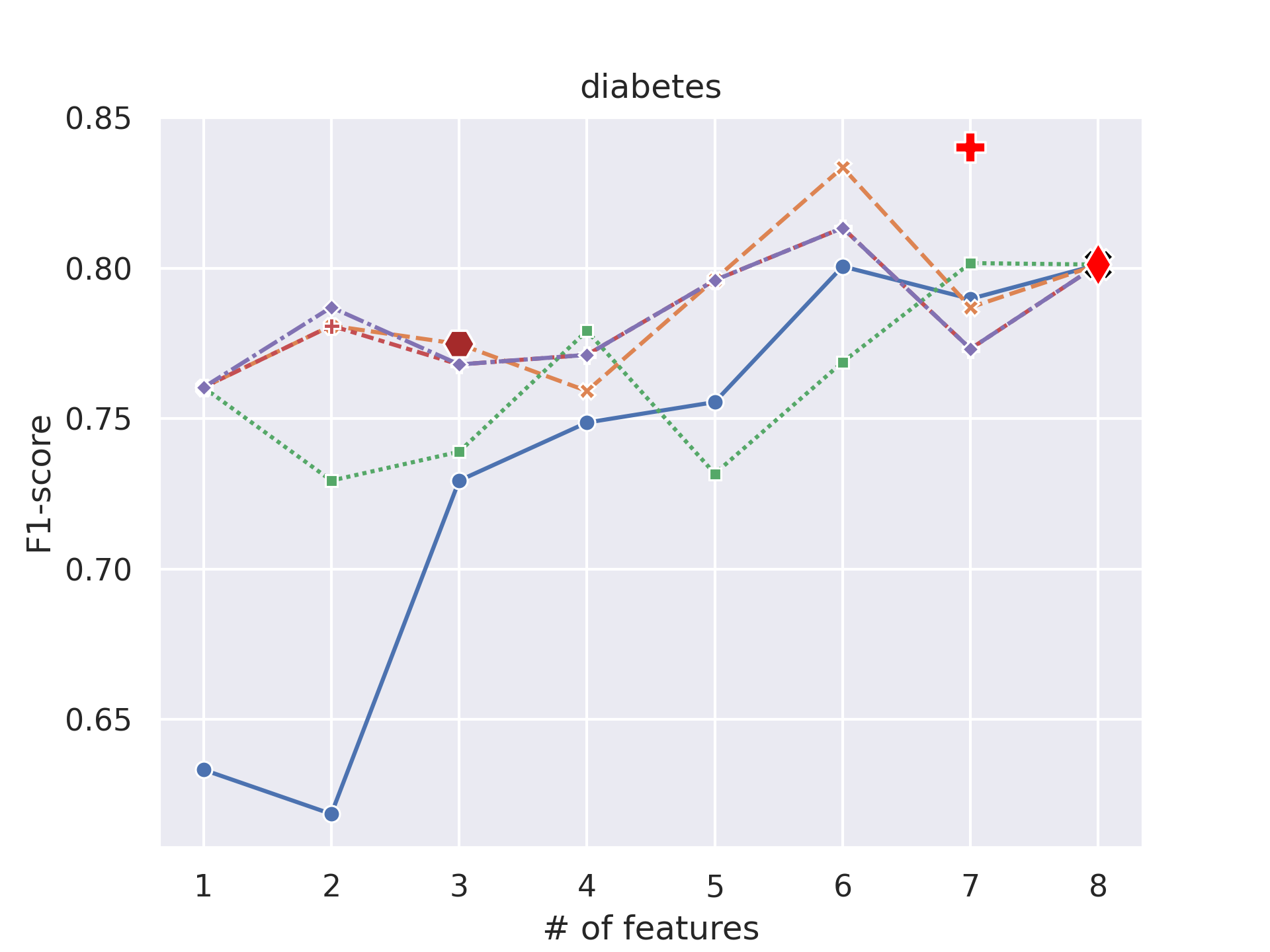}} &
\subfloat{\includegraphics[height = \height, width = \width]{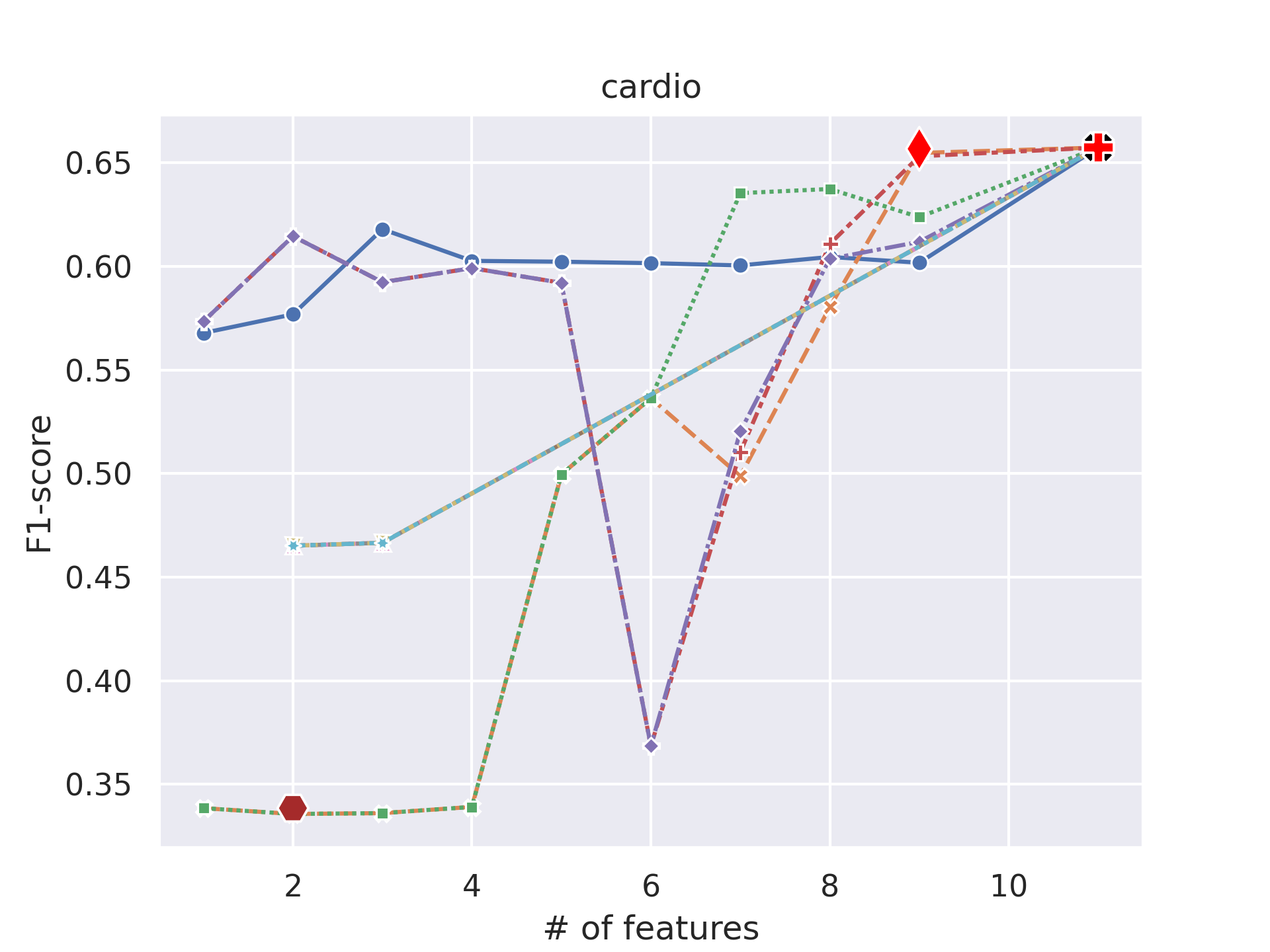}} \\[-0.52ex]

\subfloat{\includegraphics[height = \height, width = \width]{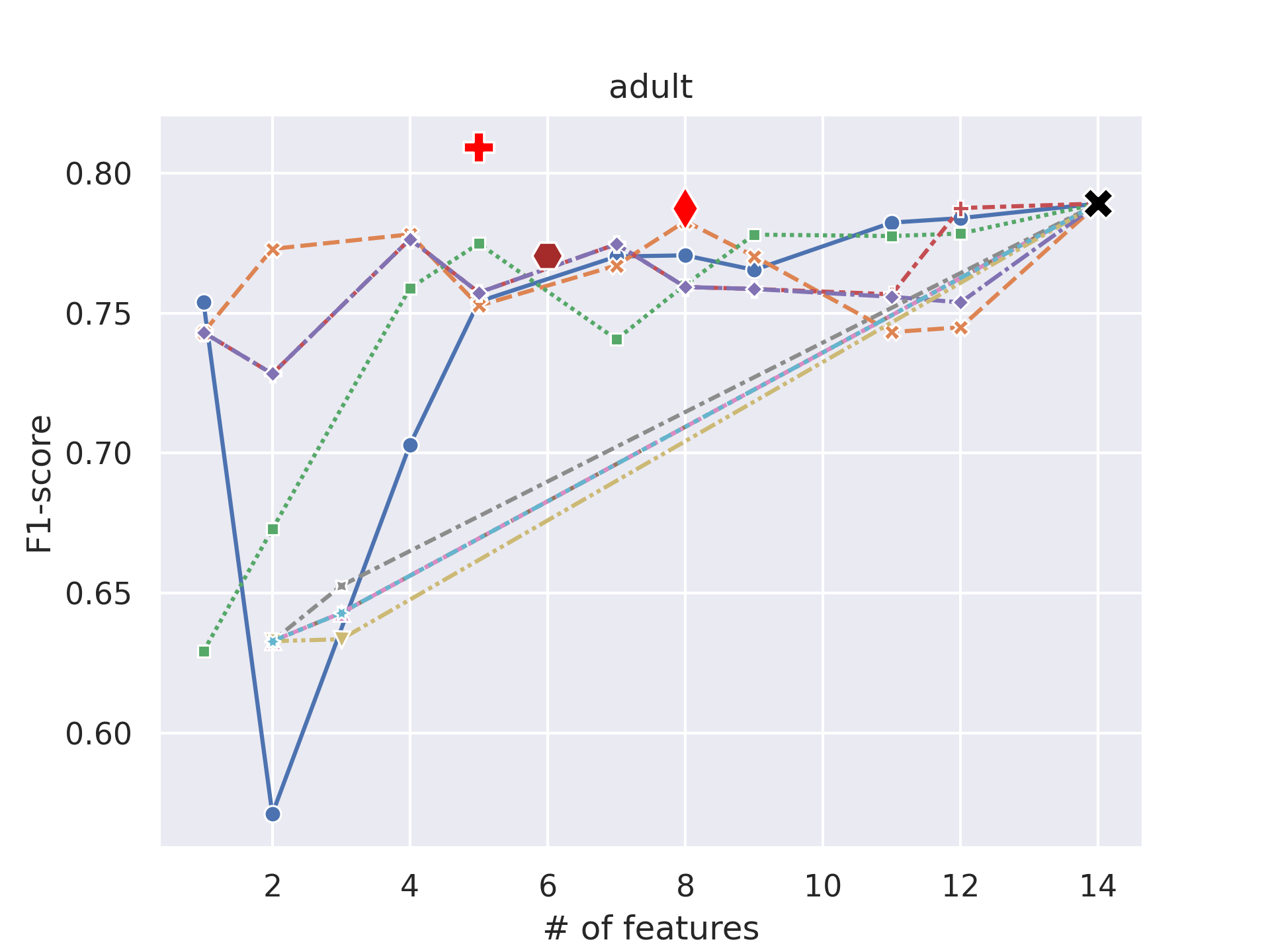}} &
\subfloat{\includegraphics[height = \height, width = \width]{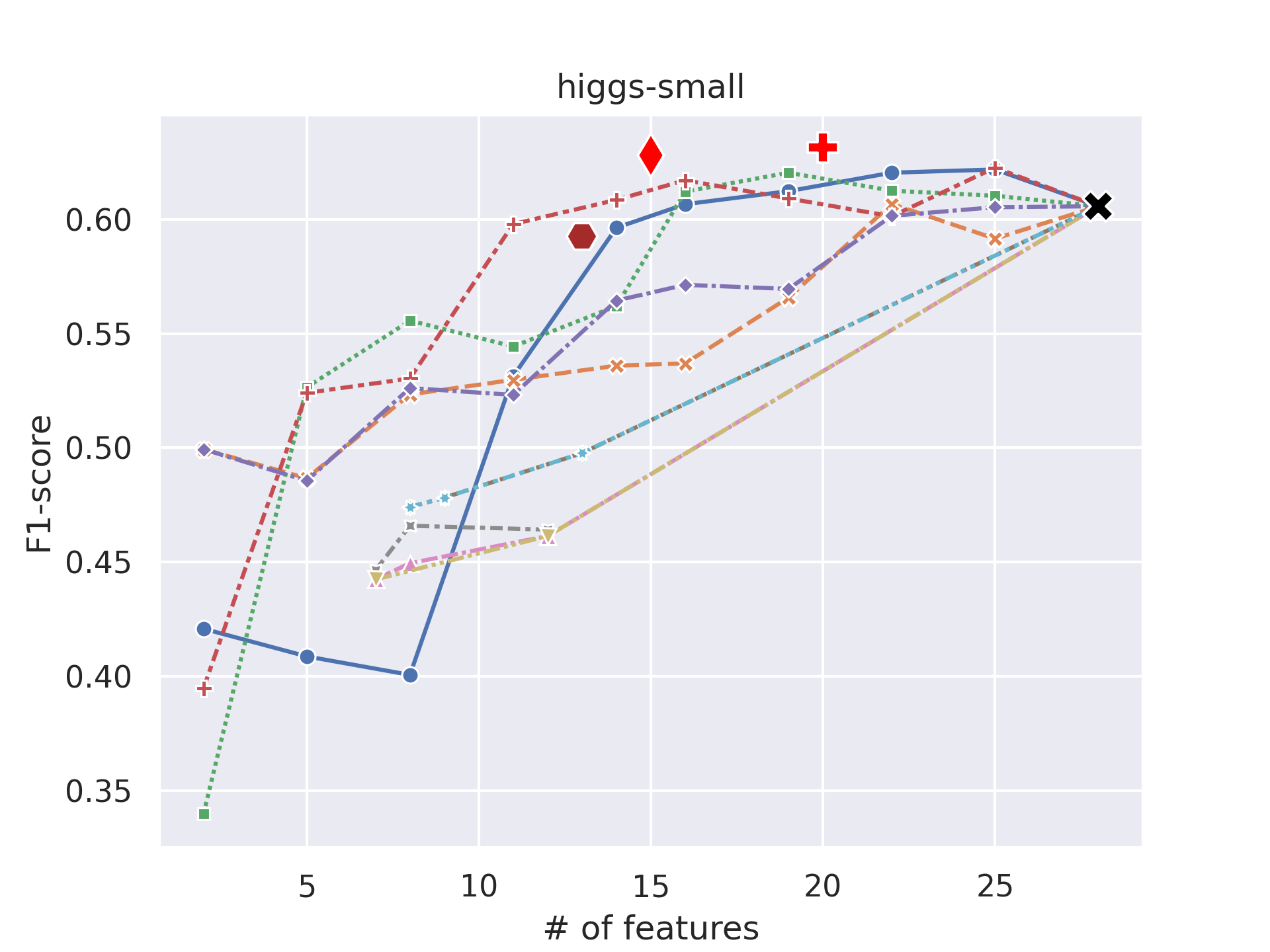}} &
\subfloat{\includegraphics[height = \height, width = \width]{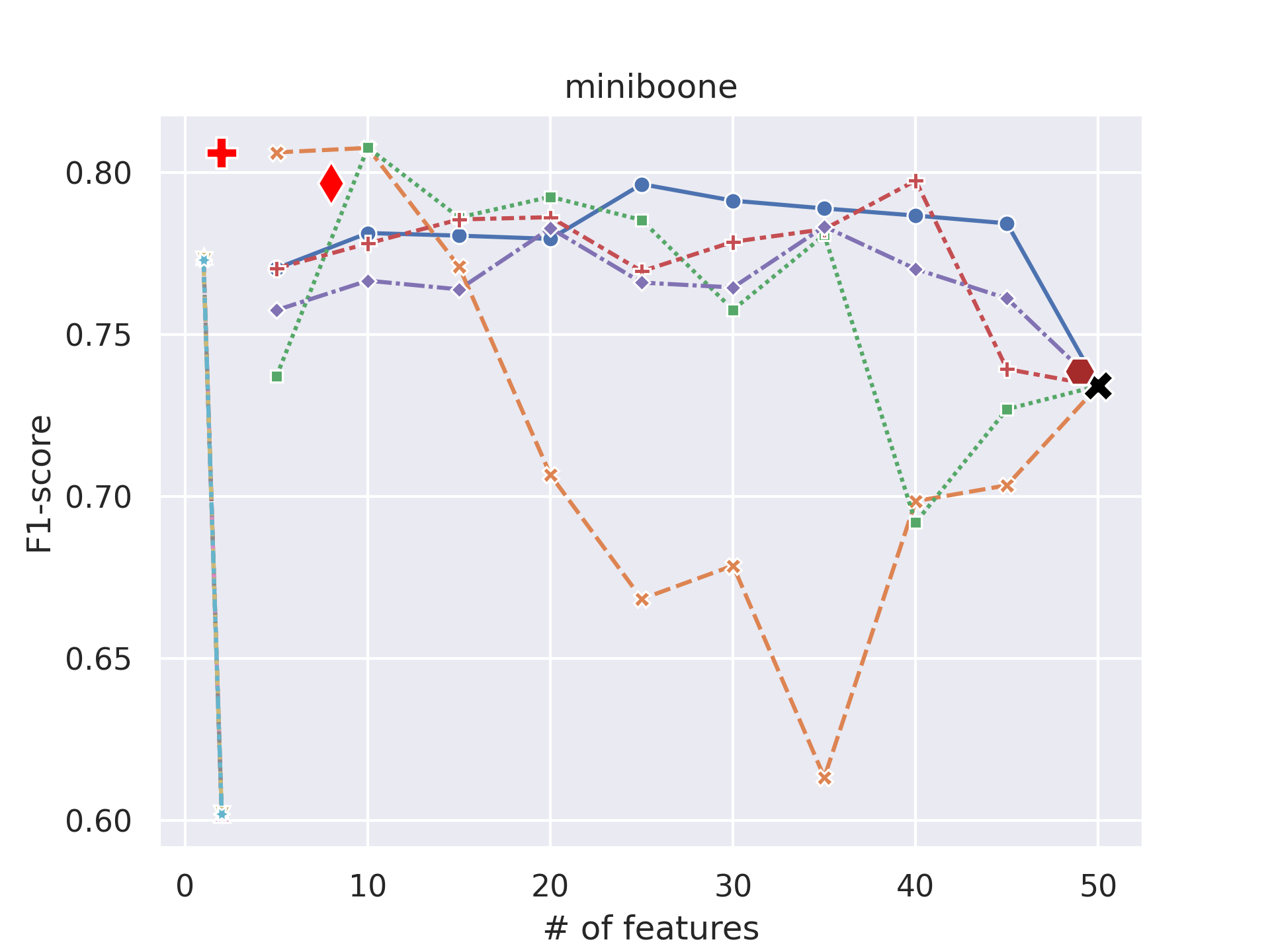}} \\[-0.52ex]

\subfloat{\includegraphics[height = \height, width = \width]{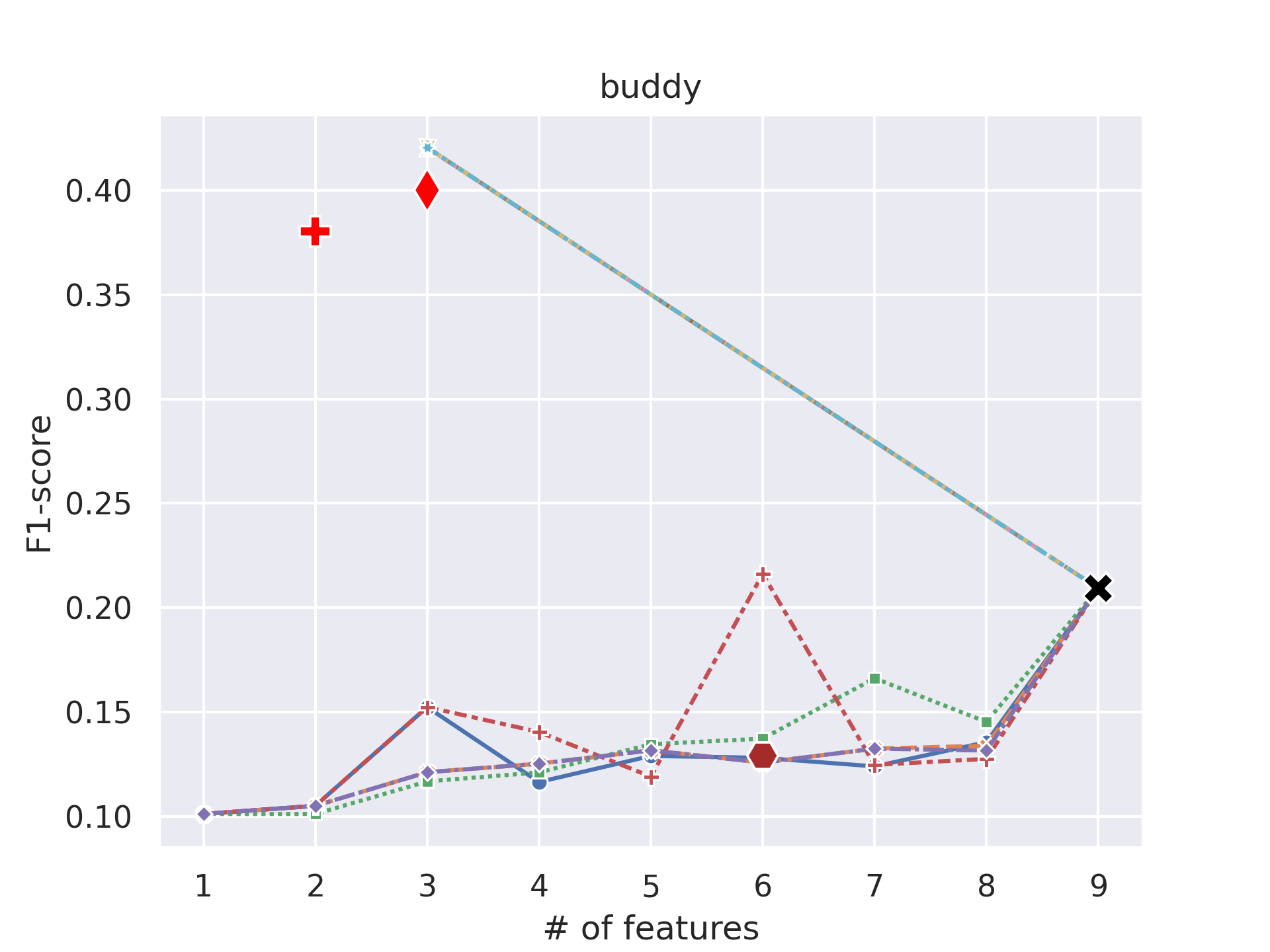}} &
\subfloat{\includegraphics[height = \height, width = \width]{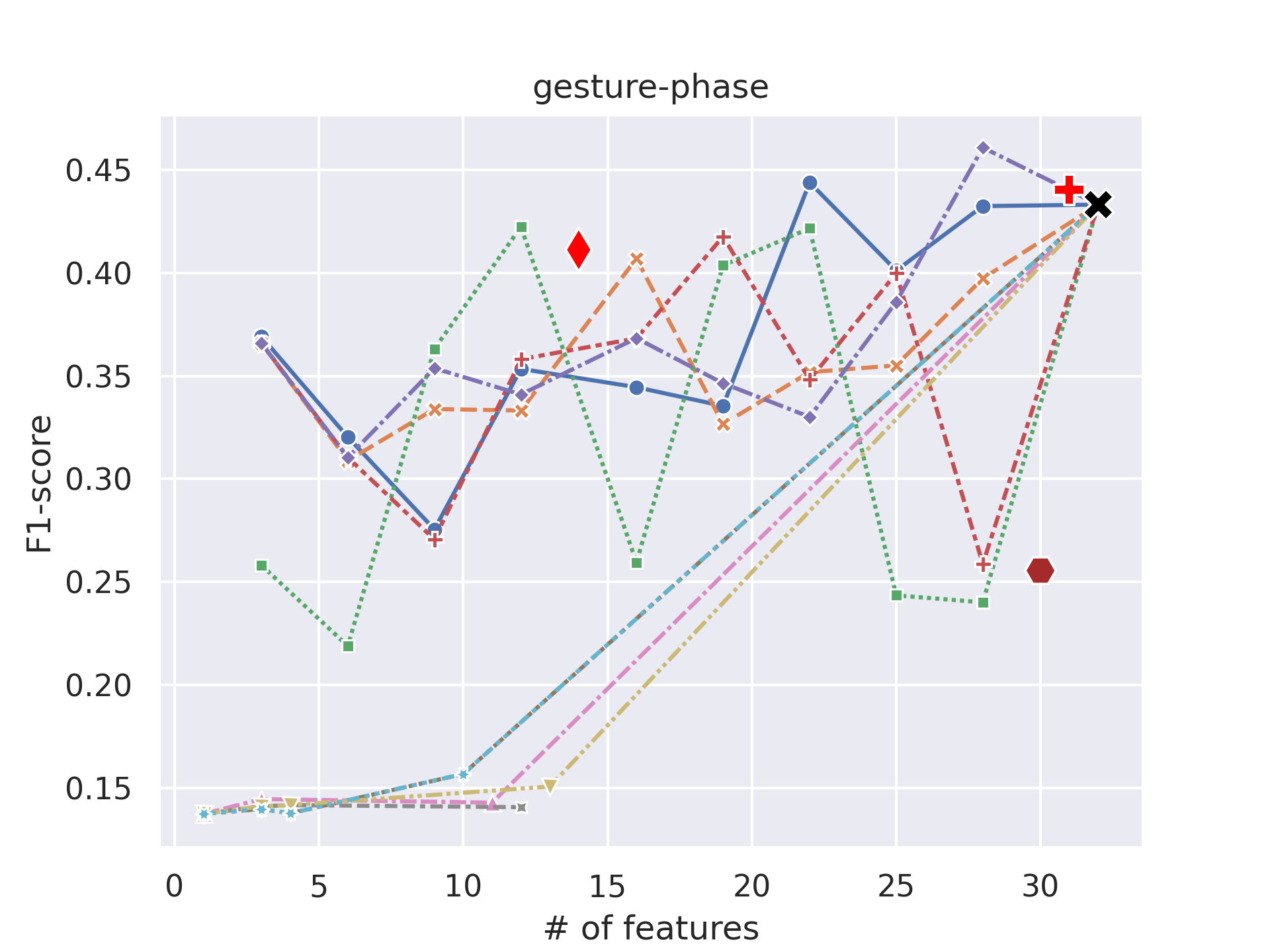}} &
\subfloat{\includegraphics[height = \height, width = \width]{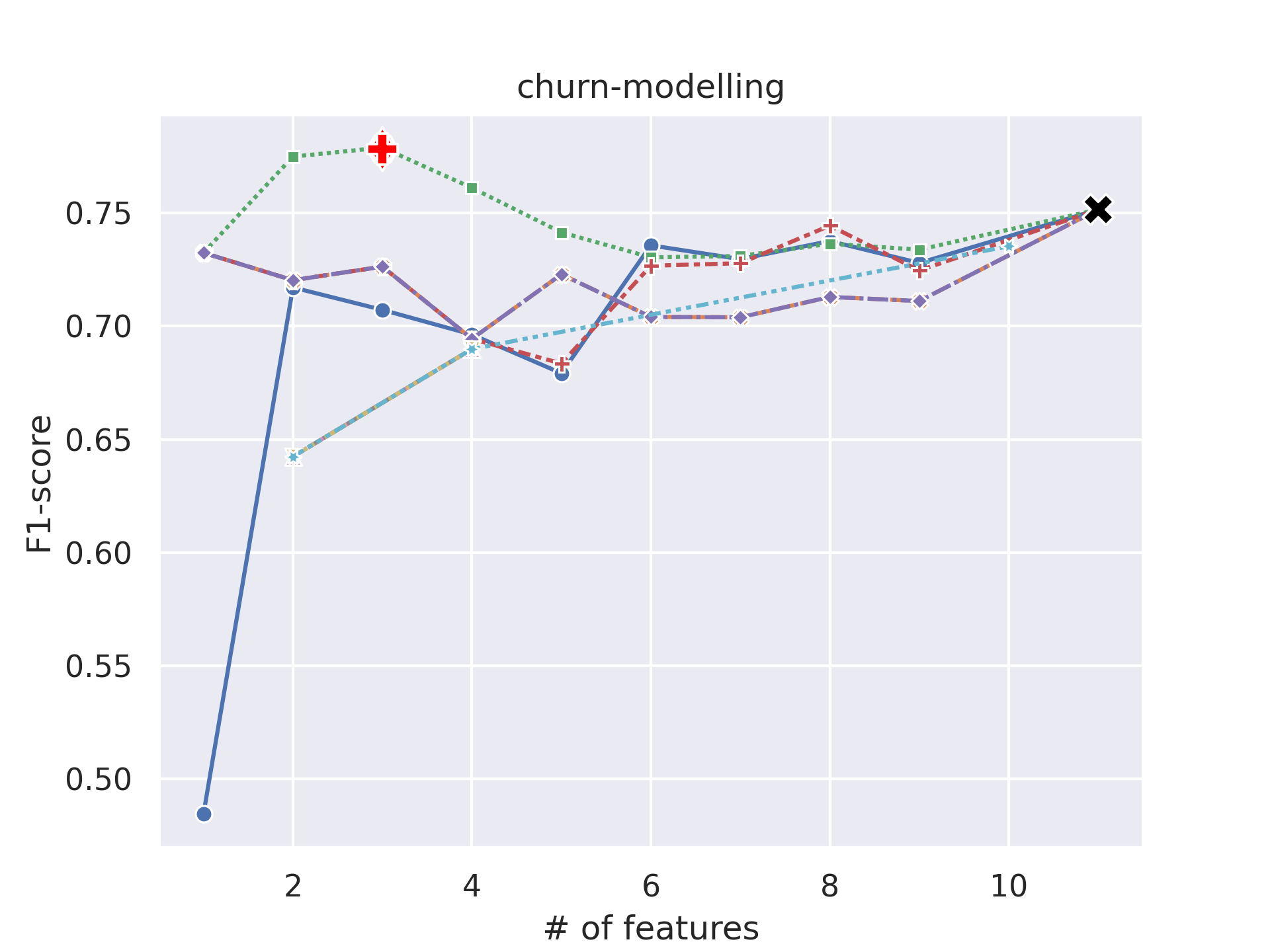}} \\[-0.52ex]

\subfloat{\includegraphics[height = \height, width = \width]{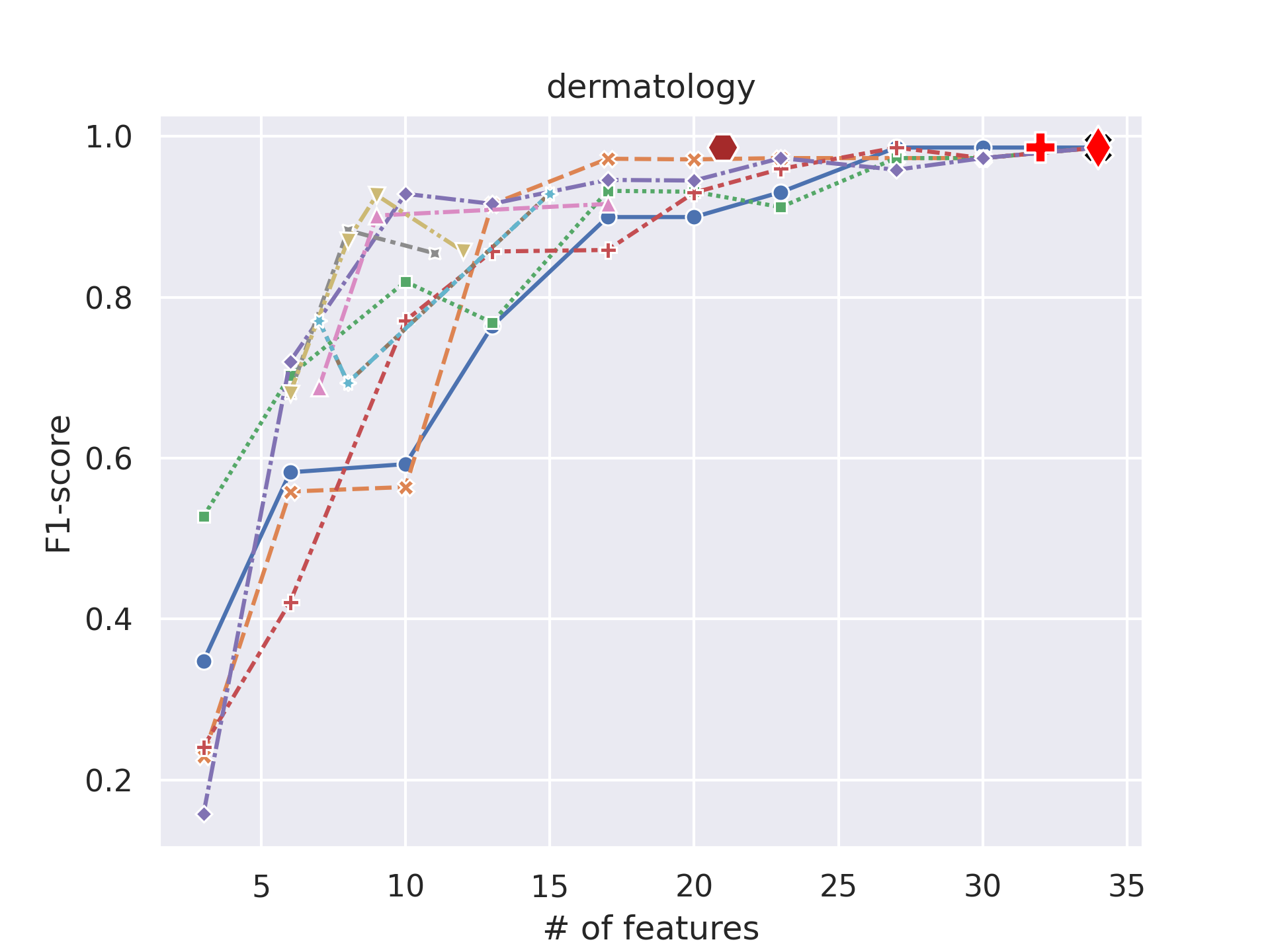}} &
\subfloat{\includegraphics[height = \height, width = \width]{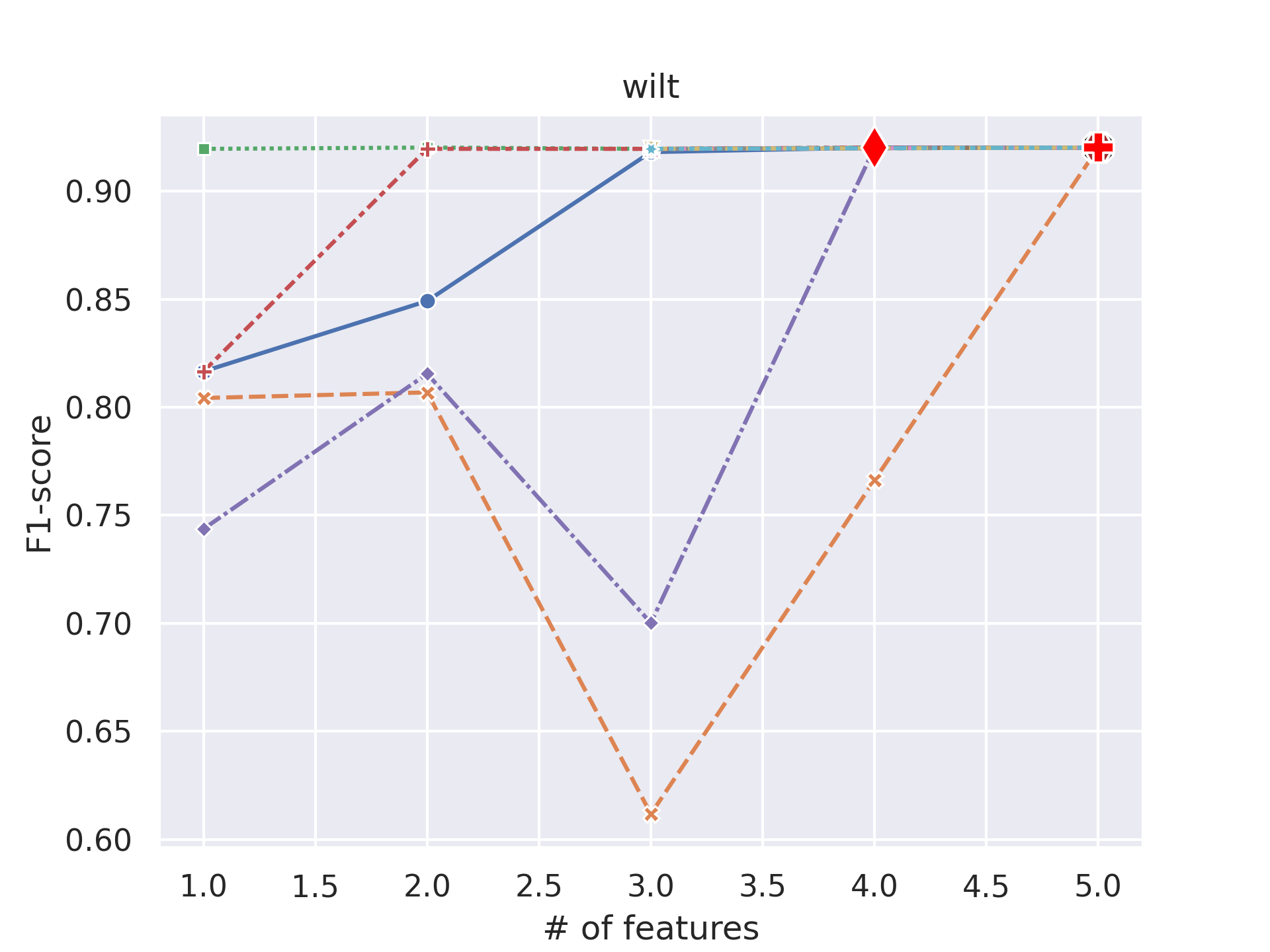}} &
 \subfloat{\includegraphics[height = \height, width = \width]{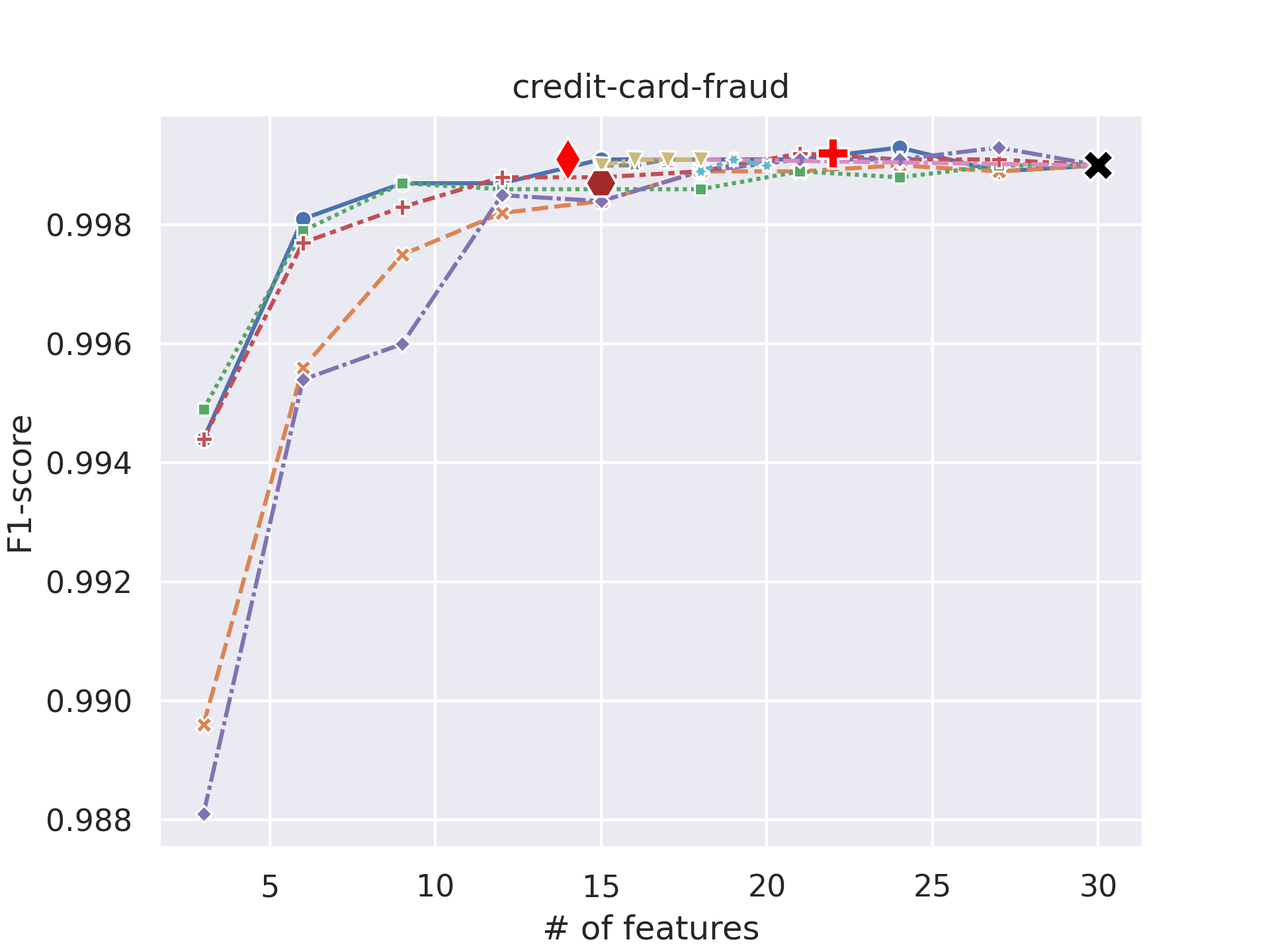}} \\[0.5ex]

\end{tabular}

}

\caption[Comparison of neural-causal vs. classical causal and non-causal feature selection on real-world datasets.]{Comparison of neural-causal vs. classical causal and non-causal feature-selection methods on 11 real-world classification datasets. Performance is measured using F1-score. For the classical feature-selection baselines, we generate performance curves by varying the sparsity level (i.e., the number of selected features), whereas neural-causal methods produce a single operating point corresponding to their learned sparse structure. The neural-causal method `causal-sk' consistently attains the highest performance for its discovered sparsity level, lying on the Pareto-optimal frontier across datasets. Notably, some baselines achieve higher performance when allowed to select more features, suggesting that the underlying causal structure may not be fully recovered.
\textbf{Diabetes, higgs-small, and credit-card-fraud are identified as datasets with natural experiments since causal-sk achieves the highest score (cf. Table~\ref{tab:real_world_cls_exp}).}
}
\label{fig:real_world_cls_exp}
  
\end{figure*}

\subsection{Real-world datasets}

We conduct experiments on eleven public datasets from \href{https://www.kaggle.com}{Kaggle} and UCI repository \citep{UCIPolicy}, widely used in the literature \citep{Zhao2021CTAB-GAN:Synthesizing,Gorishniy2021RevisitingData,KotelnikovHSE2022TabDDPM:Models}. These datasets vary in size, features, and classes, providing a robust evaluation. Dataset details are provided in Table~\ref{tab:stats}. 

We split the data into training, validation, and test sets (64:16:20 split). The training set is used to discover the causal graph and train the downstream classifiers. The validation set is used for hyperparameter tuning. The test set is only used to evaluate classifier performance. The dataset features are normalized to the range $[0,1]$. We address class imbalance by oversampling the underrepresented classes in the training data \citep{Rahman2013AddressingDatasets}.

For each dataset, we determine the optimal MLP hyperparameters by maximizing the F1-score on the val set using all features (``all-feat''). These optimal hyperparams are then used for all feature-selection methods. We fix the random seed while training the classifiers, ensuring that methods discovering the same Markov-blanket have the same accuracy (cf. Table~\ref{tab:real_world_cls_exp}), allowing for a fair comparison.

We perform causal discovery with observational ($O$) and interventional ($I_{SK}$) configurations, treating the class node as the intervened node, similar to the synthetic dataset experiment. DCDI hyperparameters are tuned for each dataset on the validation set, by optimizing the classifier's F1-score.

\subsubsection{Results}
 
The results are shown in Fig.~\ref{fig:real_world_cls_exp}, which plots F1-score as a function of the discovered sparsity level. We observe that the neural causal method  `causal-sk' consistently lies on the Pareto-optimal frontier, achieving the highest predictive performance for a given level of sparsity. These results suggest that  `causal-sk' is an effective feature-selection method, balancing predictive accuracy and model sparsity.

We further conduct a 4-fold cross-validation on the training set, comparing the best-sparse feature set found by each method. Table~\ref{tab:real_world_cls_exp} shows that the neural causal methods are competitive with other baselines, achieving high scores on 7 out of 11 datasets. 
We find that on these real-world datasets, \emph{DCDI using the soft-known intervention (causal-sk) configuration outperforms all other (observational) baselines on Diabetes, higgs-small, and credit-card-fraud datasets. This suggests the presence of natural experiments in these datasets}.

\begin{table*}[!ht]
  \caption[F1-scores of the best-sparse feature set found by each method on real-world datasets.]{
  F1-scores in percentage (mean \& std) of the best-sparse feature set found by each method (cf. Fig~\ref{fig:real_world_cls_exp}) evaluated on the test set using 4-fold cross-validation on the training set. The neural-causal methods (causal-sk \& causal-obs) show competitive performance against the baselines, achieving high scores on 7 out of 11 datasets. The soft-known (causal-sk) configuration generally performs better than the observational (causal-obs) configuration.
  \textbf{Diabetes, higgs-small, and credit-card-fraud are identified as datasets with natural experiments since causal-sk performs best.}
  \emph{We fix the random seed when training downstream classifiers to ensure that any two methods discovering the same feature set yield identical accuracy, enabling fair comparison.}
  Note, we exclude some low-performing baselines for space constraints 
  (see Appendix Table~\ref{tab:real_world_cls_exp_full} for the full comparison).
  }
  \label{tab:real_world_cls_exp}
  \centering

    \resizebox{\textwidth}{!}{

  \begin{tabular}{l|c|cccc|ccc|c}
    \toprule

     \multirow{2}{*}{Datasets} & & \multicolumn{4}{c}{Non-causal} & \multicolumn{3}{c}{Causal-observation} & \multicolumn{1}{c}{Causal-interventional} \\
    \cmidrule{3-6} \cmidrule{7-9} \cmidrule{10-10}\\[-\normalbaselineskip]
    
     & all-feat & chi-sq & spearman & mi & boruta & STMB & MMMB & \textbf{causal-obs} & \textbf{causal-sk} \\

    \midrule
    Diabetes & 79.06 $\pm$ 1.13 & 79.06 $\pm$ 1.13 & 76.18 $\pm$ 1.32 & 80.26 $\pm$ 1.02 & 77.79 $\pm$ 1.31 & 79.06 $\pm$ 1.13 & 79.06 $\pm$ 1.13 & 79.06 $\pm$ 1.13 & \textcolor{blue}{\textbf{81.00 $\pm$ 2.05}} \\
    \rowcolor{lightgray!40}
    Higgs-small & 61.76 $\pm$ 0.78 & 61.63 $\pm$ 1.40 & 60.10 $\pm$ 0.56 & 60.90 $\pm$ 0.39 & 60.21 $\pm$ 0.78 & 61.76 $\pm$ 0.78 & 61.76 $\pm$ 0.78 & 60.60 $\pm$ 0.88 & \textcolor{blue}{\textbf{62.13 $\pm$ 0.91}} \\
    Miniboone & 70.05 $\pm$ 2.28 & 78.14 $\pm$ 0.71 & \textcolor{blue}{\textbf{80.72 $\pm$ 0.07}} & 79.37 $\pm$ 0.65 & 75.97 $\pm$ 2.04 & 76.82 $\pm$ 0.20 & 76.82 $\pm$ 0.20 & 79.98 $\pm$ 0.61 & 80.58 $\pm$ 0.04 \\
    \rowcolor{lightgray!40}
    Wilt & \textbf{92.02 $\pm$ 0.00} & 91.93 $\pm$ 0.02 & \textbf{92.02 $\pm$ 0.00} & \textbf{92.02 $\pm$ 0.00} & \textbf{92.02 $\pm$ 0.00} & \textbf{92.02 $\pm$ 0.00} & \textbf{92.02 $\pm$ 0.00} & \textbf{92.02 $\pm$ 0.00} & \textbf{92.02 $\pm$ 0.00} \\
    Cardio & 33.58 $\pm$ 0.01 & 33.58 $\pm$ 0.01 & 33.58 $\pm$ 0.01 & 33.58 $\pm$ 0.01 & 33.43 $\pm$ 0.28 & 33.58 $\pm$ 0.01 & 33.58 $\pm$ 0.01 & \textcolor{blue}{\textbf{33.75 $\pm$ 0.24}} & 33.58 $\pm$ 0.01 \\
    \rowcolor{lightgray!40}
    Gesture-phase & 42.20 $\pm$ 2.68 & \textcolor{blue}{\textbf{43.77 $\pm$ 1.88}} & 42.20 $\pm$ 2.68 & 42.20 $\pm$ 2.68 & 37.62 $\pm$ 1.69 & 42.20 $\pm$ 2.68 & 42.20 $\pm$ 2.68 & 41.43 $\pm$ 1.63 & 39.97 $\pm$ 1.57 \\
    Dermatology & 97.27 $\pm$ 0.98 & 97.27 $\pm$ 0.98 & 97.27 $\pm$ 0.98 & 97.27 $\pm$ 0.98 & \textcolor{blue}{\textbf{97.94 $\pm$ 0.68}} & 91.63 $\pm$ 0.04 & 93.03 $\pm$ 1.98 & 97.27 $\pm$ 0.98 & 97.59 $\pm$ 1.14 \\
    \rowcolor{lightgray!40}
    Adult & \textbf{78.50 $\pm$ 1.13} & \textbf{78.50 $\pm$ 1.13} & \textbf{78.50 $\pm$ 1.13} & \textbf{78.50 $\pm$ 1.13} & 76.01 $\pm$ 0.43 & \textbf{78.50 $\pm$ 1.13} & \textbf{78.50 $\pm$ 1.13} & 76.63 $\pm$ 1.67 & 70.96 $\pm$ 4.80 \\
    Buddy & 14.64 $\pm$ 1.01 & 14.64 $\pm$ 1.01 & 14.64 $\pm$ 1.01 & 14.64 $\pm$ 1.01 & 13.44 $\pm$ 1.27 & 39.23 $\pm$ 4.72 & 39.23 $\pm$ 4.72 & \textcolor{blue}{\textbf{39.66 $\pm$ 0.91}} & 37.82 $\pm$ 0.86 \\
    \rowcolor{lightgray!40}
    Churn-modelling & 74.18 $\pm$ 0.27 & 74.18 $\pm$ 0.27 & 74.18 $\pm$ 0.27 & 77.71 $\pm$ 0.46 & \textbf{78.98 $\pm$ 0.94} & 71.58 $\pm$ 1.84 & 71.58 $\pm$ 1.84 & \textbf{78.98 $\pm$ 0.94} & \textbf{78.98 $\pm$ 0.94} \\
    Credit-card-fraud & 99.90 $\pm$ 0.01 & 99.90 $\pm$ 0.01 & 99.90 $\pm$ 0.01 & 99.89 $\pm$ 0.02 & 99.89 $\pm$ 0.01 & 99.90 $\pm$ 0.02 & 99.90 $\pm$ 0.02 & 99.88 $\pm$ 0.03 & \textcolor{blue}{\textbf{99.91 $\pm$ 0.01}} \\
    \bottomrule
  \end{tabular}

  } 

\end{table*}

\begin{table}[h]
  \caption[Broute-force search results on real-world datasets]{
  Broute-force search results. 
  We perform a brute-force search, evaluating all possible feature set combinations for datasets with 11 or fewer features, on the original train-val-test split. We rank the feature sets found by the `casual-sk' method based on F1-score and provide the edit-distance of the Markov-blanket found by `causal-sk' against the best performing Markov-blanket. Our observations show that `causal-sk' recovers the ideal Markov-blanket set with a max error of 2. The feature set ranks in the top 16\% or better (except on the Buddy dataset). On the Diabetes \& Cardio dataset, it recovers the ideal feature set out of 255 \& 2047 possibilities respectively, and on the Churn Modelling dataset, it ranks 54th out of 2047 possibilities.
  }
  \label{tab:brute_force_results}
\centering
\begin{tabular}{l|cc}
\toprule
Dataset           & Classifier Rank $\downarrow$ & $\bf ED_{MB} \downarrow$  \\
\midrule

Wilt & 5 / 31 & 2 \\
Diabetes & \textcolor{blue}{1} / 255 & \textcolor{blue}{0} \\
Buddy & 188 / 511 & 2 \\
Churn Modelling & 55 / 2047 & 2 \\
Cardio & \textcolor{blue}{1} / 2047 & \textcolor{blue}{0} \\

\bottomrule
\end{tabular}
\end{table}

\subsubsection{Discussion}
\begin{figure}[!h]
  \centering
    \centering
    \includegraphics[width=0.55\columnwidth]{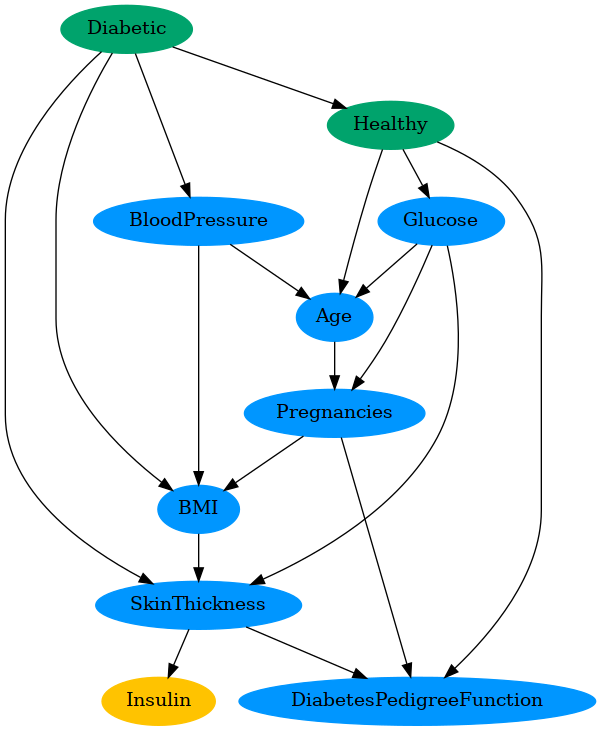}
  \caption[The DAG discovered by the causal-sk for the Diabetes dataset.]{
  The DAG discovered by \texttt{causal-sk} for the Diabetes dataset. Notably, the \emph{Insulin} feature is excluded from the Markov blanket of the class node. Although this may seem counterintuitive, an exhaustive evaluation of all 255 feature subsets confirms that the optimal feature set also excludes \emph{Insulin} (Table~\ref{tab:brute_force_results}). This finding is consistent with the original study \citep{Bennett1976EpidemiologicIndians}, which notes that insulin responses vary across glucose-tolerance groups, suggesting that insulin may not be a consistently reliable predictor of diabetes status. This highlights the ability of \texttt{causal-sk} to disregard unreliable predictors in favor of features with stronger causal relevance, unlike all evaluated baseline methods, which selected \emph{Insulin}.
  }
  \label{fig:diabetes_dag}
\end{figure}

For the diabetes dataset, the causal-sk model excludes the insulin (2-Hour serum insulin (mu U/ml)) feature, deeming it unreliable 
(refer to Fig~\ref{fig:diabetes_dag} for the discovered DAG). 
Because this was counterintuitive, we investigated this result further. A brute-force evaluation of all 255 feature combinations using the 8 features of the diabetes dataset revealed that the best feature set indeed excludes the insulin feature 
(see Table~\ref{tab:brute_force_results}), 
corroborating the findings of the causal-sk model. Further investigation led us to a 1976 paper by the original authors \citep{Bennett1976EpidemiologicIndians} who published this diabetes dataset, which states, ``Insulin responses among Pima Indians show different patterns according to the level of glucose tolerance.''
Although insulin is often a good predictive metric for diabetes, it is not always reliable and can be misleading in certain cases \citep{Bennett1976EpidemiologicIndians,Hupfeld2015TypeHistory,Jameson2015Endocrinology:Pediatric,Metter2008GlucosePrediction}. Therefore, while a greedy approach might include the insulin feature, a causal approach would learn to ignore it.

\section{Conclusion}
We demonstrated we can uncover natural experiments in real-world datasets when the interventional causal feature selection method fares better than other observational feature selection methods. We systematically evaluated this claim on controlled synthetic graphs and a suite of real-world datasets. We also showed the utility of neural-causal discovery algorithm DCDI as a viable feature-selection method for classification on real-world datasets, finding that DCDI gives the best results in soft-known intervention configuration.
Our findings suggest that real-world datasets may indeed contain natural experiments, and accounting for these interventions can improve performance, prompting a reevaluation of the inherent nature of existing datasets and the optimal strategies for their utilization.

\section{Acknowledgments}
The authors gratefully acknowledge Prof. Peter Spirtes, Gokul Swamy, Bennett P. deBoisblanc, Tom Fox, Ricardo Luis Rodriguez, Jia Shi, Mihir Prabhudesai, and Neehar Peri for their valuable discussions and contributions to this work. This research was supported in part by the Carnegie Mellon University (CMU) Center for Machine Learning and Health (CMLH) Translational Fellowship in Digital Health. The authors also thank the Weights \& Biases team \cite{wandb} for providing their experiment tracking platform.

\bibliographystyle{unsrtnat}
\bibliography{references,main}

\pagebreak

\appendix 
%














\tableofcontents

\pagebreak
\listoffigures
\listoftables

\pagebreak

\section{Additional Details for the main paper}

\begin{figure}[!ht]
  \centering
    \includegraphics[width=0.5\columnwidth]{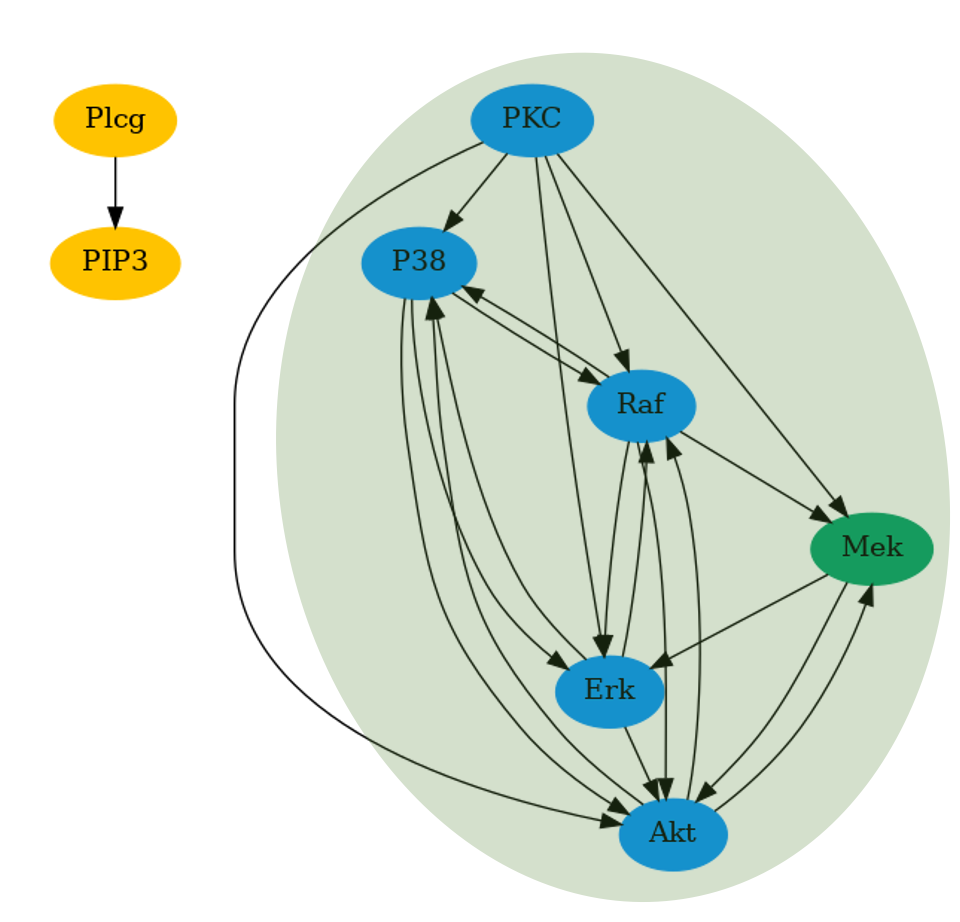}
  
  \caption[Marginalized Mixed Acyclic Graph (MAG) of the Sachs DAG]{The marginalized Mixed Acyclic Graph (MAG) of the Sachs DAG graph after marginalizing the nodes PKA, PIP2, and Jnk. In our hidden confounder experiments, we treat the Mek variable (the green node) as the target class to be predicted when measuring downstream classification accuracy with hidden confounders PKA, PIP2, and Jnk. It’s new Markov-blanket (the green region) now includes  PKC, P38, Raf, P38, and Erk. Hidden confounders (the grey nodes) are PKA, PIP2, and Jnk.}
  \label{fig:sachs_skip_graph}

\end{figure}

\begin{table}[!ht]
  \caption{The main hyper-parameters tuned}
  \label{tab:hyperparams}
  \centering
  
  
  \begin{tabular}{l|c}
    \toprule
    
    Hyperparameter     &    Search space \\
    
    \midrule
    \multicolumn{2}{c}{Causal discovery (DCDI) } \\  
    \midrule

    Learning rate   &  LogUniform$[10^{-6}, 0.5]$ \\
    Regularization co-efficient $\lambda$   &  Uniform$[0.1, 2]$ \\
    \# of iterations   &  $[500000]$ \\
    Model & DeepSigmoidalFlow (DSF) \\
    
    \midrule
    \multicolumn{2}{c}{Classifier (MLP) } \\  
    \midrule

    Learning rate   &  LogUniform$[10^{-5}, 0.01]$ \\
    \# of layers   &  UniformInt$[1, 6]$ \\
    Width of layers   &  $2^{Int[3, 10]}$ \\
    Epochs   &  Int$[200, 1000]$ \\
    Batch size   &  Int$[32, 256, 2048]$ \\
    
    \bottomrule
  \end{tabular}

  
\end{table}

\begin{table}[!ht]
  \caption[F1-scores of the best-sparse feature set found by each method]{
    F1-scores in percentage (mean \& std) of the best-sparse feature set found by each method 
      (cf. Fig~2) 
    evaluated on the test set using 4-fold cross-validation on the training set. The neural-causal methods (causal-sk \& causal-obs) show competitive performance against the baselines, achieving high scores on 7 out of 11 datasets. The soft-known (causal-sk) configuration generally performs better than the observational (causal-obs) configuration.
  \textbf{Diabetes, higgs-small, and credit-card-fraud are identified as datasets with natural experiments since causal-sk performs best.}
  \emph{We fix the random seed when training downstream classifiers to ensure that any two methods discovering the same feature set yield identical accuracy, enabling fair comparison.}
  Note, this table shows the full comparison with all baseline methods expanding on the main paper Table~3.
  }
  \label{tab:real_world_cls_exp_full}
  \centering

  \resizebox{\textwidth}{!}{

  \begin{tabular}{l|c|cccccc|cccccc|c}
    \toprule

     \multirow{2}{*}{Datasets} & & \multicolumn{6}{c}{Non-causal} & \multicolumn{6}{c}{Causal-observation} & \multicolumn{1}{c}{Causal-interventional} \\
    \cmidrule{3-8} \cmidrule{9-13} \cmidrule{14-15}\\[-\normalbaselineskip]
    
     & all-feat & chi-sq & spearman & mi & anova & pearson & boruta & LRH & STMB & HITON-MB & MMMB & IAMB & \textbf{causal-obs} & \textbf{causal-sk} \\

    \midrule
   
    Diabetes & 79.06 $\pm$ 1.13 & 79.06 $\pm$ 1.13 & 76.18 $\pm$ 1.32 & 80.26 $\pm$ 1.02 & 76.80 $\pm$ 2.67 & 76.80 $\pm$ 2.67 & 77.79 $\pm$ 1.31 & 79.06 $\pm$ 1.13 & 79.06 $\pm$ 1.13 & 79.06 $\pm$ 1.13 & 79.06 $\pm$ 1.13 & 79.06 $\pm$ 1.13 & 79.06 $\pm$ 1.13 & \textcolor{blue}{\textbf{81.00 $\pm$ 2.05}} \\
    \rowcolor{lightgray!40}
    Higgs-small & 61.76 $\pm$ 0.78 & 61.63 $\pm$ 1.40 & 60.10 $\pm$ 0.56 & 60.90 $\pm$ 0.39 & 61.63 $\pm$ 1.40 & 61.76 $\pm$ 0.78 & 60.21 $\pm$ 0.78 & 61.76 $\pm$ 0.78 & 61.76 $\pm$ 0.78 & 47.41 $\pm$ 1.01 & 61.76 $\pm$ 0.78 & 61.76 $\pm$ 0.78 & 60.60 $\pm$ 0.88 & \textcolor{blue}{\textbf{62.13 $\pm$ 0.91}} \\
    Miniboone & 70.05 $\pm$ 2.28 & 78.14 $\pm$ 0.71 & \textcolor{blue}{\textbf{80.72 $\pm$ 0.07}} & 79.37 $\pm$ 0.65 & 79.28 $\pm$ 0.76 & 77.59 $\pm$ 0.49 & 75.97 $\pm$ 2.04 & 76.82 $\pm$ 0.20 & 76.82 $\pm$ 0.20 & 76.82 $\pm$ 0.20 & 76.82 $\pm$ 0.20 & 76.82 $\pm$ 0.20 & 79.98 $\pm$ 0.61 & 80.58 $\pm$ 0.04 \\
    \rowcolor{lightgray!40}
    Wilt & \textbf{92.02 $\pm$ 0.00} & 91.93 $\pm$ 0.02 & \textbf{92.02 $\pm$ 0.00} & \textbf{92.02 $\pm$ 0.00} & 91.93 $\pm$ 0.02 & \textbf{92.02 $\pm$ 0.00} & \textbf{92.02 $\pm$ 0.00} & \textbf{92.02 $\pm$ 0.00} & \textbf{92.02 $\pm$ 0.00} & \textbf{92.02 $\pm$ 0.00} & \textbf{92.02 $\pm$ 0.00} & \textbf{92.02 $\pm$ 0.00} & \textbf{92.02 $\pm$ 0.00} & \textbf{92.02 $\pm$ 0.00} \\
    Cardio & 33.58 $\pm$ 0.01 & 33.58 $\pm$ 0.01 & 33.58 $\pm$ 0.01 & 33.58 $\pm$ 0.01 & 33.58 $\pm$ 0.01 & 33.58 $\pm$ 0.01 & 33.43 $\pm$ 0.28 & 33.58 $\pm$ 0.01 & 33.58 $\pm$ 0.01 & 33.58 $\pm$ 0.01 & 33.58 $\pm$ 0.01 & 33.58 $\pm$ 0.01 & \textcolor{blue}{\textbf{33.75 $\pm$ 0.24}} & 33.58 $\pm$ 0.01 \\
    \rowcolor{lightgray!40}
    Gesture-phase & 42.20 $\pm$ 2.68 & \textcolor{blue}{\textbf{43.77 $\pm$ 1.88}} & 42.20 $\pm$ 2.68 & 42.20 $\pm$ 2.68 & 42.20 $\pm$ 2.68 & 43.49 $\pm$ 1.01 & 37.62 $\pm$ 1.69 & 42.20 $\pm$ 2.68 & 42.20 $\pm$ 2.68 & 13.98 $\pm$ 0.17 & 42.20 $\pm$ 2.68 & 42.20 $\pm$ 2.68 & 41.43 $\pm$ 1.63 & 39.97 $\pm$ 1.57 \\
    Dermatology & 97.27 $\pm$ 0.98 & 97.27 $\pm$ 0.98 & 97.27 $\pm$ 0.98 & 97.27 $\pm$ 0.98 & 97.27 $\pm$ 0.98 & 97.27 $\pm$ 0.98 & \textcolor{blue}{\textbf{97.94 $\pm$ 0.68}} & 91.23 $\pm$ 0.72 & 91.63 $\pm$ 0.04 & 90.99 $\pm$ 2.52 & 93.03 $\pm$ 1.98 & 91.23 $\pm$ 0.72 & 97.27 $\pm$ 0.98 & 97.59 $\pm$ 1.14 \\
    \rowcolor{lightgray!40}
    Adult & \textbf{78.50 $\pm$ 1.13} & \textbf{78.50 $\pm$ 1.13} & \textbf{78.50 $\pm$ 1.13} & \textbf{78.50 $\pm$ 1.13} & \textbf{78.50 $\pm$ 1.13} & \textbf{78.50 $\pm$ 1.13} & 76.01 $\pm$ 0.43 & \textbf{78.50 $\pm$ 1.13} & \textbf{78.50 $\pm$ 1.13} & \textbf{78.50 $\pm$ 1.13} & \textbf{78.50 $\pm$ 1.13} & \textbf{78.50 $\pm$ 1.13} & 76.63 $\pm$ 1.67 & 70.96 $\pm$ 4.80 \\
    Buddy & 14.64 $\pm$ 1.01 & 14.64 $\pm$ 1.01 & 14.64 $\pm$ 1.01 & 14.64 $\pm$ 1.01 & 14.64 $\pm$ 1.01 & 14.64 $\pm$ 1.01 & 13.44 $\pm$ 1.27 & 39.23 $\pm$ 4.72 & 39.23 $\pm$ 4.72 & 39.23 $\pm$ 4.72 & 39.23 $\pm$ 4.72 & 39.23 $\pm$ 4.72 & \textcolor{blue}{\textbf{39.66 $\pm$ 0.91}} & 37.82 $\pm$ 0.86 \\
    \rowcolor{lightgray!40}
    Churn-modelling & 74.18 $\pm$ 0.27 & 74.18 $\pm$ 0.27 & 74.18 $\pm$ 0.27 & 77.71 $\pm$ 0.46 & 74.18 $\pm$ 0.27 & 74.18 $\pm$ 0.27 & \textbf{78.98 $\pm$ 0.94} & 71.58 $\pm$ 1.84 & 71.58 $\pm$ 1.84 & 71.58 $\pm$ 1.84 & 71.58 $\pm$ 1.84 & 73.50 $\pm$ 0.78 & \textbf{78.98 $\pm$ 0.94} & \textbf{78.98 $\pm$ 0.94} \\
    Credit-card-fraud & 99.90 $\pm$ 0.01 & 99.90 $\pm$ 0.01 & 99.90 $\pm$ 0.01 & 99.89 $\pm$ 0.02 & 99.91 $\pm$ 0.00 & 99.90 $\pm$ 0.01 & 99.89 $\pm$ 0.01 & 99.90 $\pm$ 0.01 & 99.90 $\pm$ 0.02 & 99.90 $\pm$ 0.01 & 99.90 $\pm$ 0.02 & 99.90 $\pm$ 0.01 & 99.88 $\pm$ 0.03 & \textcolor{blue}{\textbf{99.91 $\pm$ 0.01}} \\

    \bottomrule
  \end{tabular}

  } 

\end{table}


\section{Limitations}
Our paper questions the true nature of existing real-world datasets, typically assumed to be observational. Our experiments suggests that we can get better performance on certain datasets by treating them as interventional. While our evaluation is robust, it is not comprehensive, so our findings should be seen as suggestive, not definitive, evidence of the presence of interventional data in these datasets.

We find that DCDI training is highly sensitive to hyperparameter choices, often leading to non-convergence or cyclic graphs. Therefore, a comprehensive search for optimal hyperparameters is required for each dataset, limiting its usability compared to more straightforward feature selection methods.

\section{Impact Statement}
Our work highlights a new perspective on existing datasets, questioning their inherent nature and the best strategies for their use. Our findings suggest that some real-world datasets can be better modeled as interventional. This prompts a reevaluation of the nature of existing datasets and the best strategies for using them.


\section{Dataset and Codebase links}
We used public datasets and public codebase repositories for carrying out the experiments in the papers. The below Table~\ref{tab:dataset_codebase_links} shows the consolidated links for all the datasets and code repos used.

\begin{table}[!ht]
  \caption[Links to codebase and datasets]{The links to codebase repositories and public datasets used in the paper}
  \label{tab:dataset_codebase_links}
  \centering
  
  \resizebox{\columnwidth}{!}{
  
  \begin{tabular}{l|c}
    \toprule
    
    Codebase     &    Link \\
    
    \midrule
    
    All Public Datasets (sourced from TabDDPM [59] \href{https://github.com/yandex-research/tab-ddpm}{codebase})   &  \href{https://www.dropbox.com/s/rpckvcs3vx7j605/data.tar?dl=0}{https://www.dropbox.com/s/rpckvcs3vx7j605/data.tar?dl=0} \\

    DCDI Codebase & \href{https://github.com/slachapelle/dcdi}{https://github.com/slachapelle/dcdi} \\
    
    (Non-neural) Causal Feature selection baselines code & \href{https://github.com/wt-hu/pyCausalFS}{https://github.com/wt-hu/pyCausalFS} \\
    
    \bottomrule
  \end{tabular}

  } 
  
\end{table}

\section{Additional Results for the main paper}

\begin{table}[h]
  \caption[Average SHD and ED for the causal methods across the four synthetic data settings]{
  The average SHD and ED for the causal methods across the four observational and interventional settings carried out in synthetic Sachs data experiment 
  (refer Table~1). 
  We find that the Causal $I_{SK}$ has the best overall $avg-ED_{MB}$ score followed by the Causal $O$ method. This motivated us to evaluate the real-world datasets using both Causal $I_{SK}$ and Causal $O$ methods.
  }
  \label{tab:avg_scores_for_sachs_exp1}
\centering
\begin{tabular}{l|cc}
\toprule
Method           & $avg-SHD_G \downarrow$ & $\bf avg-ED_{MB} \downarrow$ \\
\midrule
Causal $O$       & 14.75  & 0.50 \\ 
Causal $I_{SK}$  & 12.75  & \textbf{0.25} \\ 
Causal $I_{HK}$  & 13.25  & 1.00 \\ 
Causal $I_{HU}$  & 12.75  & 1.00 \\ 
\bottomrule
\end{tabular}
\end{table}

\pagebreak

\begin{table}
  \caption[Optimal classifier hyper-parameters for the datasets]{The optimal classifier hyper-parameters found for the datasets using the Bayesian hyper-parameter search.}
  \label{tab:best-classifier-hyperparams}
  \centering
  
  \resizebox{\columnwidth}{!}{
  
  \begin{tabular}{|l|ccccc|}
    \toprule

    Dataset & Training Epochs & Batch Size & Width of layers & \# of layers & Learning rate \\
    \midrule
    
    Adult & 1000 & 256 & 32 & 2 & 0.003 \\
    Buddy & 1000 & 256 & 32 & 5 & 0.0005 \\
    Cardio & 1000 & 256 & 256 & 5 & 0.00001 \\
    Churn Modelling & 1000 & 256 & 256 & 6 & 0.0001 \\
    Credit Card Fraud & 1000 & 2048 & 256 & 2 & 0.0005 \\
    Dermatology & 200 & 32 & 128 & 1 & 0.01 \\
    Diabetes & 1000 & 256 & 32 & 6 & 0.00005 \\
    Gesture Phase & 1000 & 256 & 256 & 1 & 0.0001 \\
    Higgs Small & 1000 & 128 & 256 & 7 & 0.0003526 \\
    Miniboone & 1000 & 128 & 256 & 4 & 0.0001339 \\
    Wilt & 1000 & 256 & 8 & 2 & 0.001 \\
    
    \bottomrule
  \end{tabular}

  } 
  
\end{table}




\pagebreak

\section{The DAG graph plots found by the causal-sk method}


\begin{figure}[!ht]
  \centering
    \includegraphics[width=0.5\columnwidth]{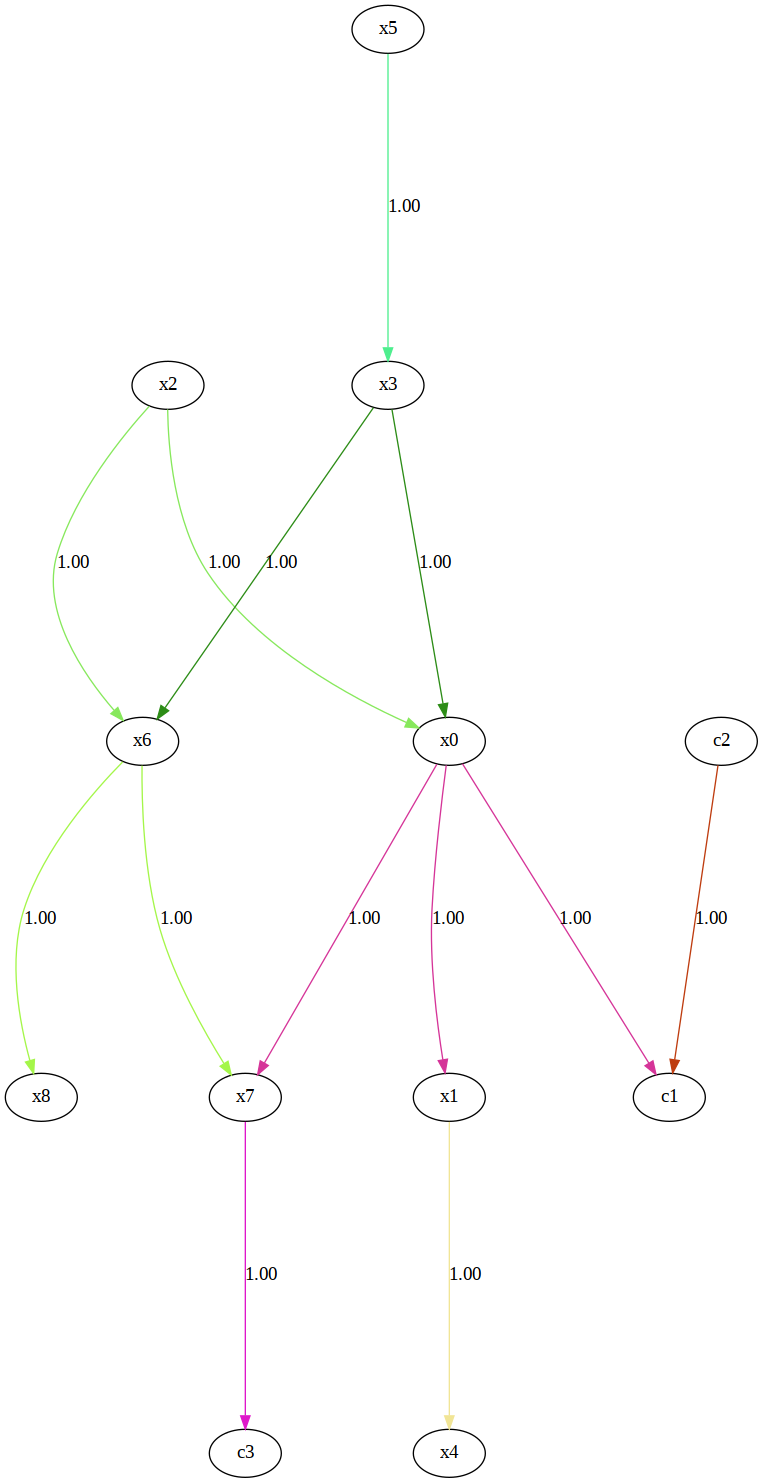}
  
  \caption{The DAG discovered by the causal-sk for the Buddy dataset.}
  \label{fig:buddy_dag}

\end{figure}

\begin{figure}[!ht]
  \centering
  \includegraphics[width=0.98\columnwidth]{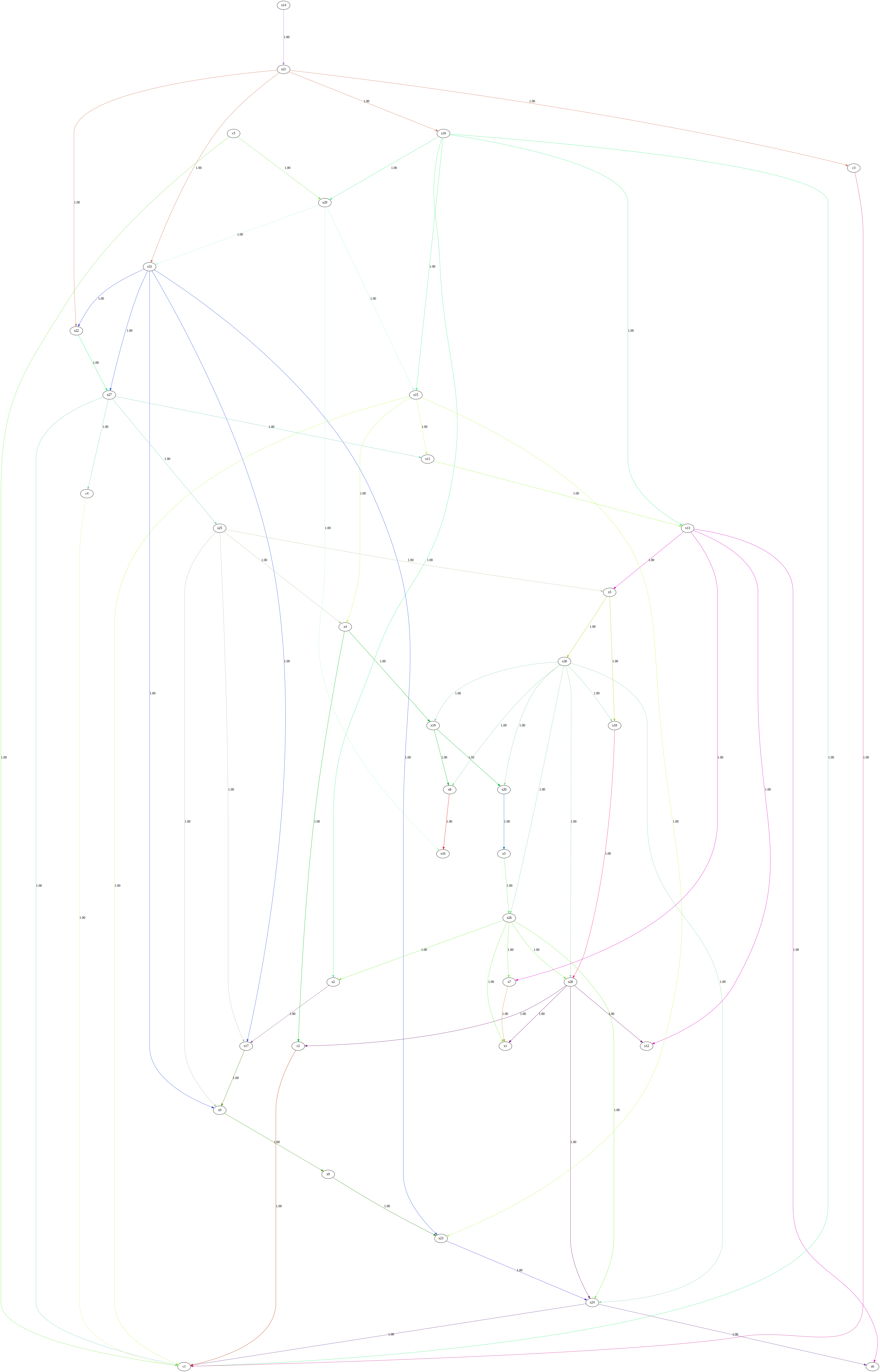}
  \caption{The DAG discovered by the causal-sk for the Gesture-phase dataset.}
  \label{fig:gesture_dag}

\end{figure}


\section{Simple medical example of causal modeling}
\label{app:simple_ex}
Let us motivate our approach with a simple example. Consider a medical dataset assessing the mortality rate of the general population being collected continuously over the years, with age as one of the features. However, in 2020 and 2021, a significant portion of the population was infected with COVID-19, representing an intervention where the coronavirus is the intervening agent. For those infected, the mortality rate is notably higher than usual. If we train a conventional ML model on this data, its prediction would be an average risk of both sub-populations combined. Consequently, such a model would be biased to overestimate the risk for healthy individuals 
(cf. Fig~1 left). 
In contrast, a causal model that is aware of the intervention would correctly predict the risk by separately accounting for the healthy and infected sub-populations. The causal model would recognize the structural break caused by the pandemic and fit two different models. 
The data generation code is attached at the end of the document.
%
\begin{figure}[h]
  \centering
    \includegraphics[width=.9\columnwidth]{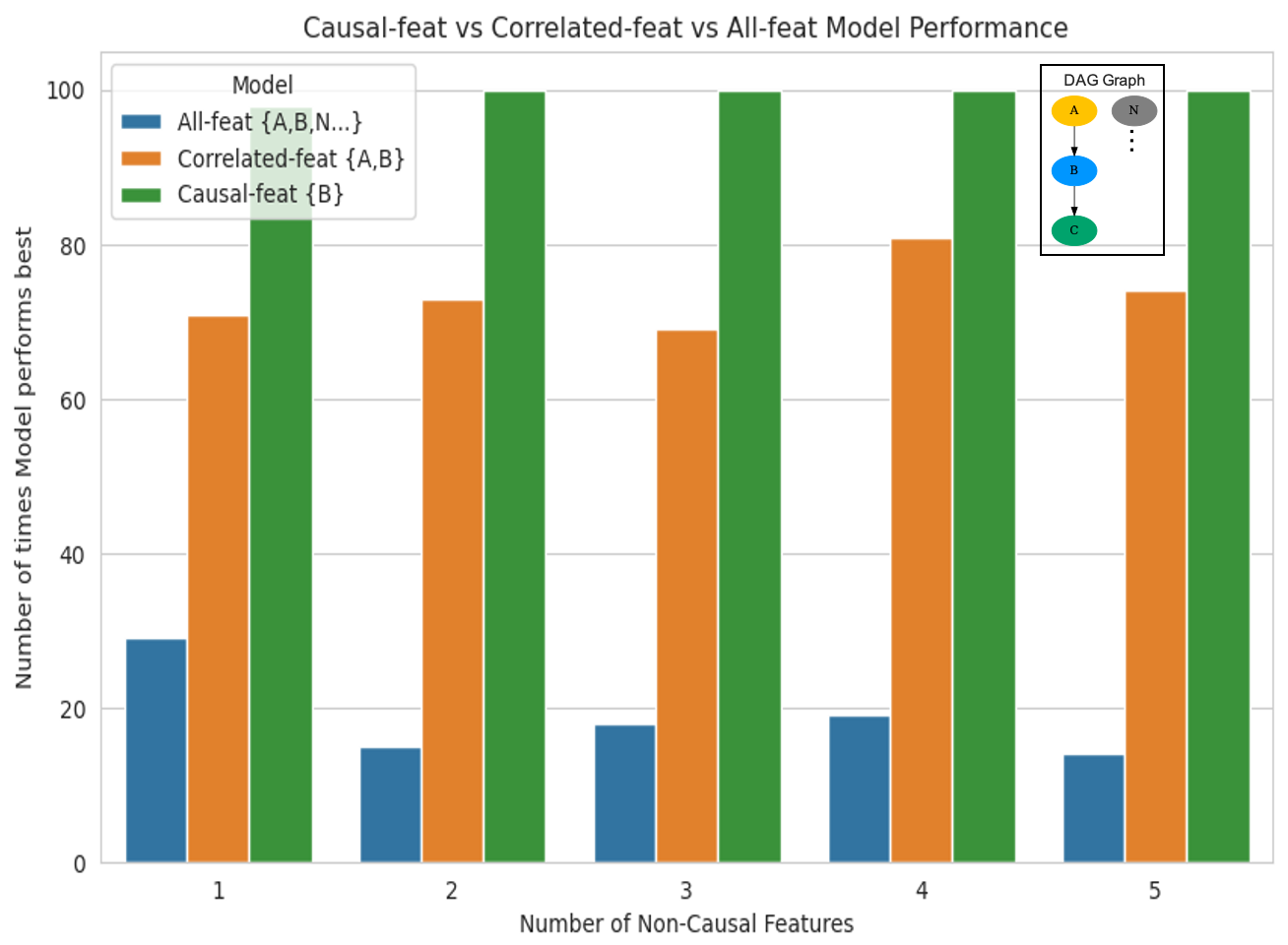}
  
  \caption[Synthetic dataset result comparing causal feature vs all-feature classifiers]{
  The figure presents the results of our analysis conducted on a synthetic dataset generated according to the Directed Acyclic Graph (DAG) shown in the inset. Three types of classifiers were trained: (1) Causal-feat utilizing the causal feature {B}, (2) Correlated-feat employing correlated features {A, B}, and (3) All-feat incorporating both correlated and non-correlated features {A, B, N...}. Through 100 random trials, we assessed the best-performing classifiers in each run, with ties equally rewarded. Notably, as the number of non-causal, non-correlated features increased, the performance of the All-feat model decreased. Intriguingly, the Causal model consistently demonstrated superior performance, highlighting the significance of feature selection strategies that prioritize causal features for optimal model performance.
  Code is attached at the end of the document.
  }
  \label{fig:syn_result_causal_vs_all_feat_clf}

\end{figure}

\pagebreak

\section{Simple demo code files}

\begin{figure}[!ht]
    \centering
    \includegraphics[width=\textwidth]{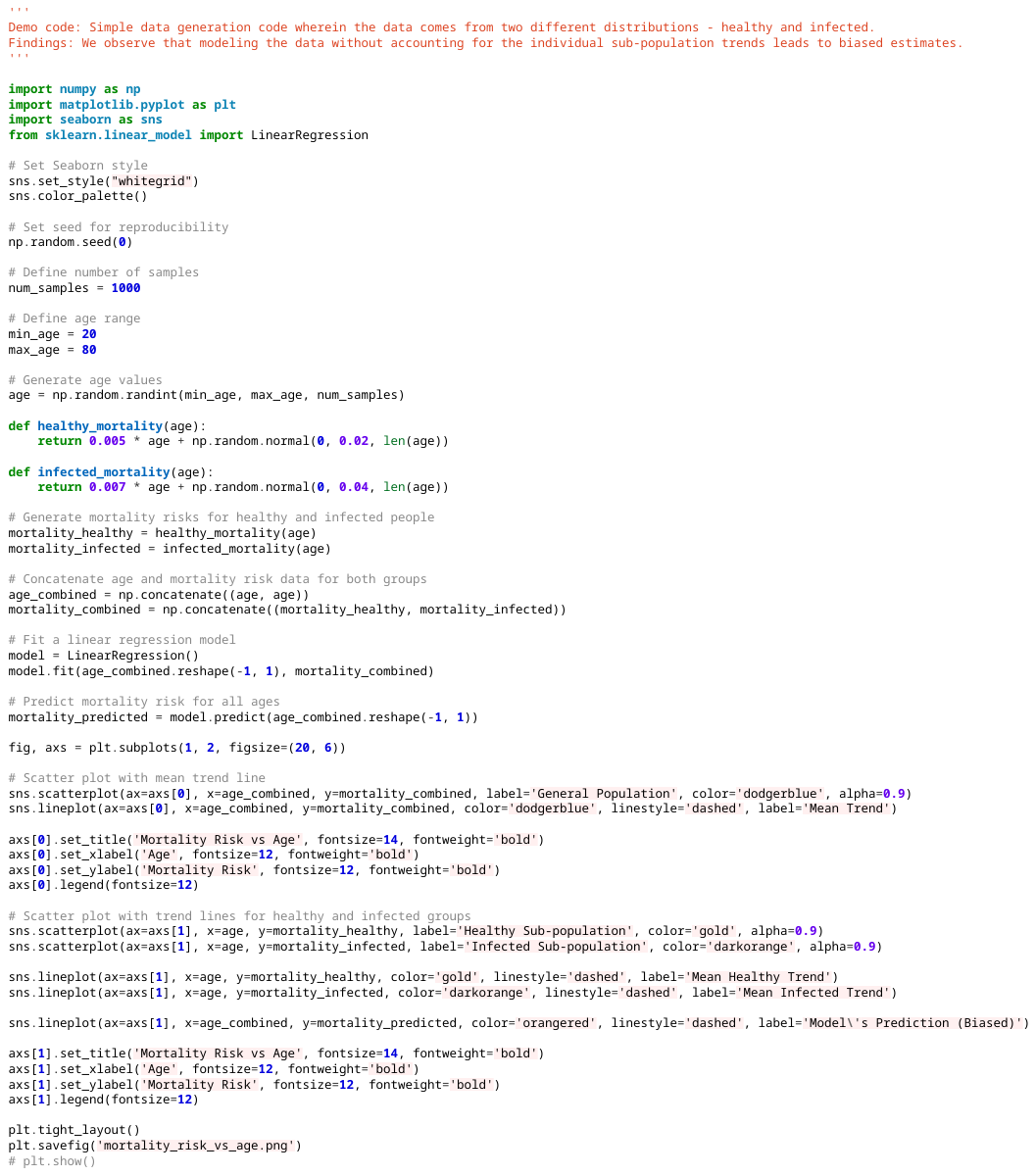}
\end{figure}

\begin{figure}[!ht]
    \centering
    \includegraphics[width=\textwidth]{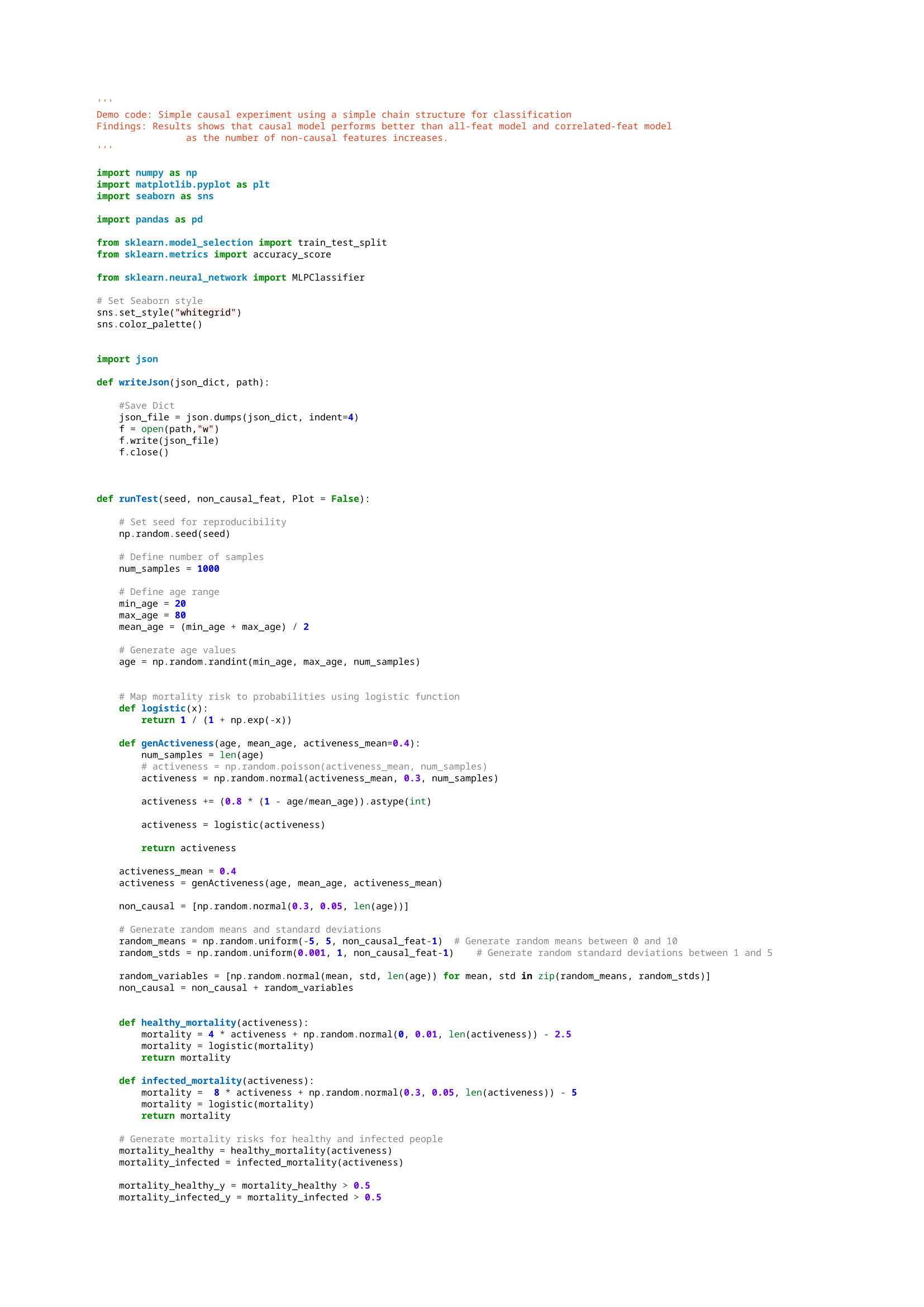}
\end{figure}

\begin{figure}[!ht]
    \centering
    \includegraphics[width=\textwidth]{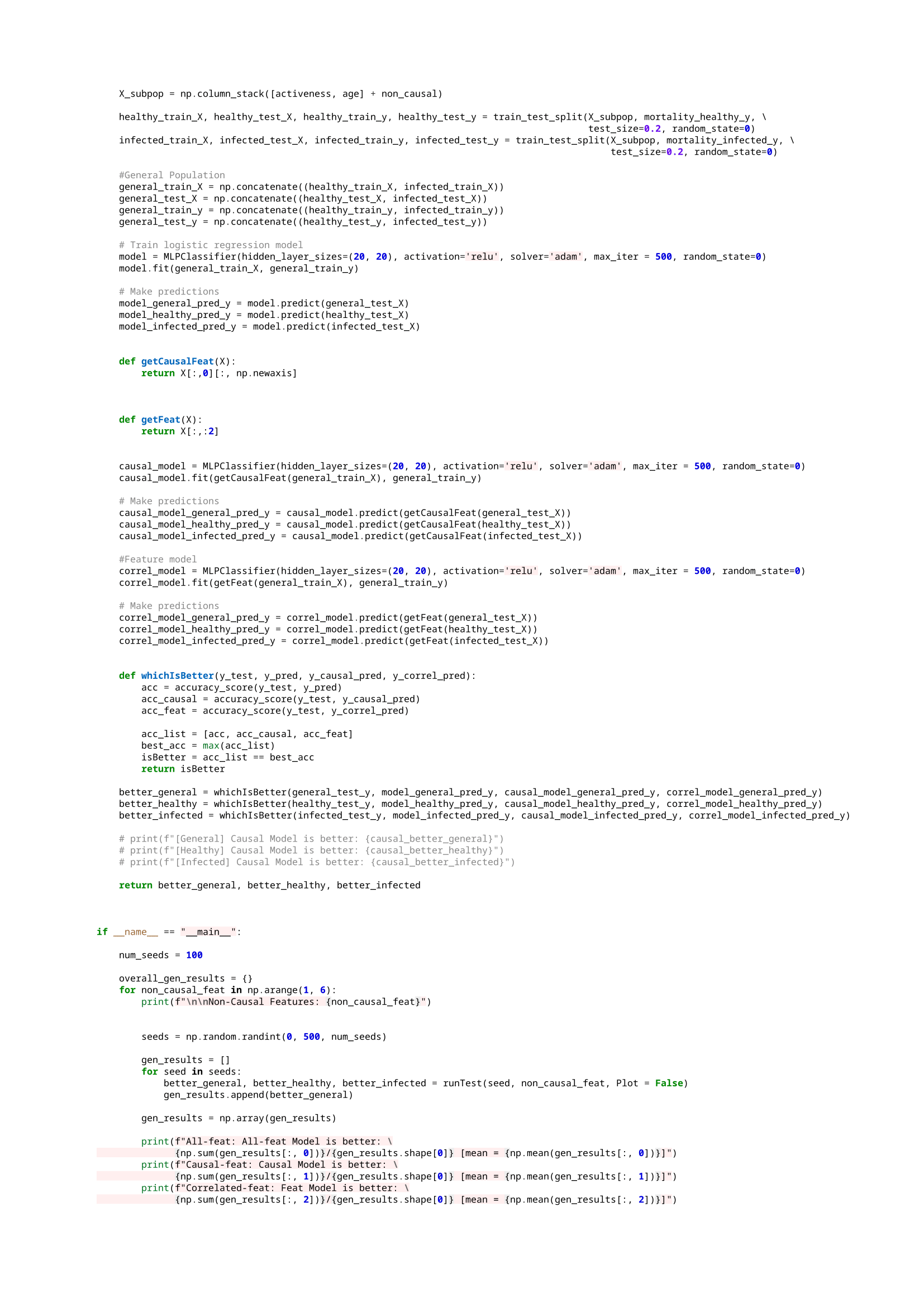}
\end{figure}

\begin{figure}[!ht]
    \centering
    \includegraphics[width=\textwidth]{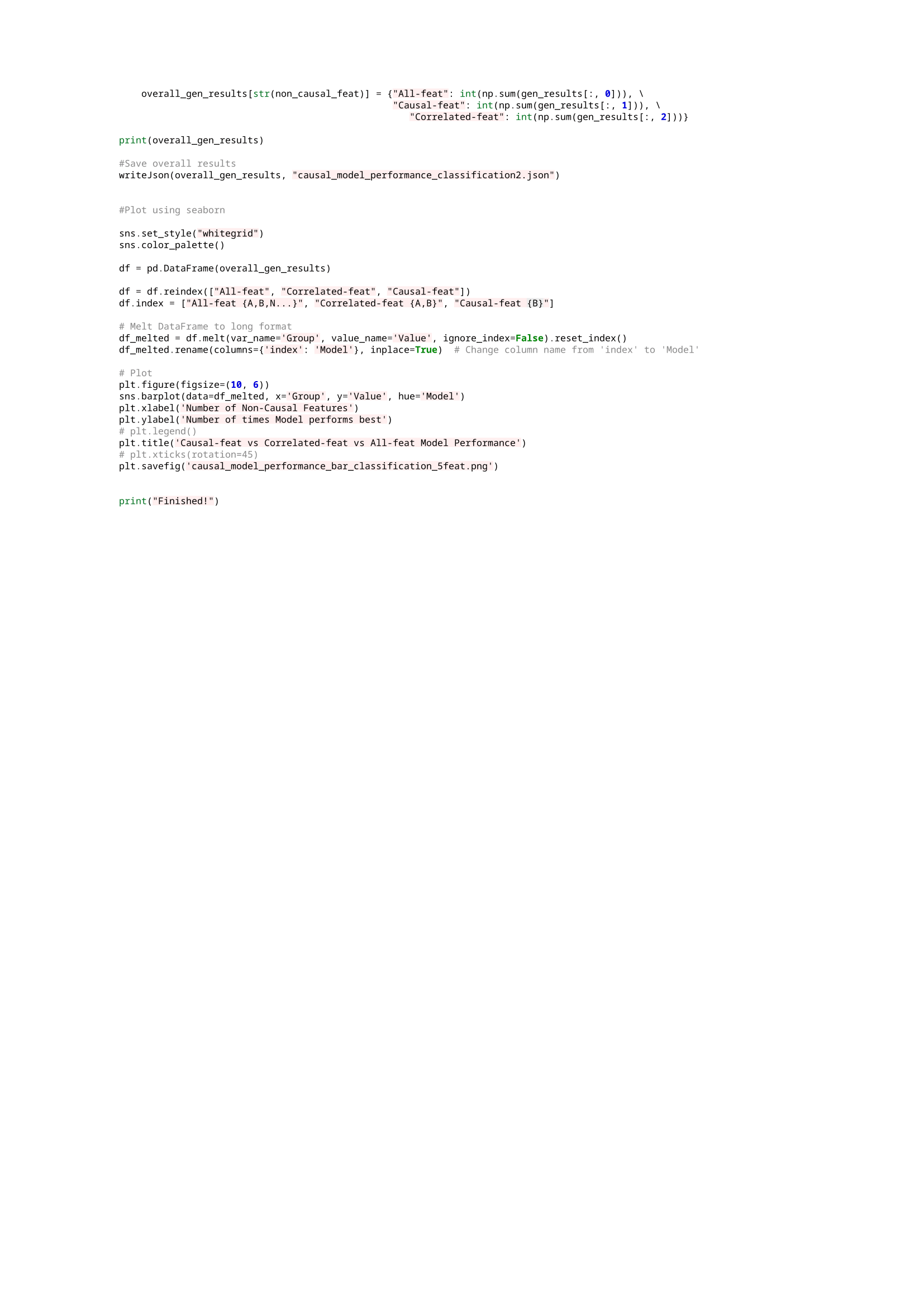}
\end{figure}


\end{document}